\documentclass[10pt]{article} 
\usepackage[preprint]{tmlr}
\usepackage{booktabs}
\usepackage{graphicx}

\usepackage{amsmath}    
\usepackage{amssymb}    
\usepackage{amsthm}     
\usepackage{algorithm}
\usepackage{algpseudocode}
\usepackage{subcaption}
\usepackage{wrapfig}

\usepackage{amsmath,amsfonts,bm}

\def\eqref#1{equation~\ref{#1}}

\def\1{\bm{1}}

\DeclareMathAlphabet{\mathsfit}{\encodingdefault}{\sfdefault}{m}{sl}
\SetMathAlphabet{\mathsfit}{bold}{\encodingdefault}{\sfdefault}{bx}{n}

\usepackage{hyperref}
\usepackage{url}

\title{Hyperflux \\ Pruning Reveals Importance}

\author{\name Barbulescu Eugen \email Barbulescu.Oc.Eugen@student.utcluj.ro\\
      \addr Department of Computer Science\\
      Technical University of Cluj-Napoca
      \AND
      \name Antonio Alexoaie \email Alexoaie.Ov.Antonio@student.utcluj.ro\\
      \addr Department of Computer Science\\
      Technical University of Cluj-Napoca
      \AND
      \name Lucian Busoniu \email lucian.busoniu@aut.ro\\
      \addr Department of Automation\\
      Technical University of Cluj-Napoca}

\def\month{MM} 
\def\year{YYYY}
\def\openreview{\url{https://openreview.net/forum?id=XXXX}}

\begin{document}

\maketitle

\begin{abstract}

Network pruning is used to reduce inference latency and power consumption in large neural networks. However, most methods focus on empirical results at the expense of understanding the pruning process. We introduce Hyperflux, a novel \(L_0\) method which models pruning as a continuously evolving system determined by flux, the gradient response to a weight's removal, and pressure, a global regularization driving weights toward pruning. By exploiting this model, Hyperflux's pruning behavior becomes understandable at both microscopic (weight regrowth/pruning) and macroscopic (sparsity convergence, etc.) levels. We also introduce a novel pressure scheduler that reliably targets desired sparsities. Hyperflux achieves competitive results with ResNet-50, VGG-19 and DeiT-T/S on CIFAR-10, CIFAR-100 and ImageNet datasets.

\end{abstract}

\section{Introduction}
\label{section:introduction}

Overparameterization has become the norm in modern deep learning to achieve state-of-the-art performance \citep{neyshabur2019overparametrization, allenzhu2019convergence, li2018intrinsic}. Despite clear benefits for training, this practice also increases computational and memory costs, complicating deployment on resource-constrained devices such as edge hardware, IoT platforms, and autonomous robots \citep{shi2016edge, li2019edgeai}. Recent theoretical and empirical findings suggest that sparse subnetworks extracted from large dense models can match or exceed the accuracy of their dense counterparts \citep{frankle19lottery, zhou19deconstructing, ma2021sanity, lee2019snip, dejorge2021progressiveforce, cho2024pdp, wang2023ntksap, frantar2024scaling, wang2023loft} and even outperform smaller dense models of equal size \citep{ramanujan2020hidden, li2020training, zhu2018topruneornot}. These results have created interest in network pruning as a strategy to identify minimal, high-performing subnetworks.

Pruning has a rich history \citep{lecun1989optimal, skeletonization1988, thimm1995evaluating} and continues to prove valuable for real-time applications \citep{han2016deepcompression, park2017faster, wang2019hardwareaware}. Recent methods have significantly advanced the field by resorting to a variety of strategies and heuristics, from magnitude pruning, gradient methods, and Hessian-based criteria \citep{han2015learning, han2016deepcompression, lecun1992second, singh2020efficient, bellec2018deeprewiring, frankle19lottery} to dynamic pruning approaches \citep{liu2020dst,cho2024pdp, savarese2020winningcontinuous, kusupati2020soft, wortsman2019discovering} or combinations thereof \citep{liu2021sparse, evci2020rigging}. However, the strong interdependence between weights remains a challenge \citep{jin2020weight, anthropic2024scaling, lee2019snip, dejorge2021progressiveforce, christos2017bayesian}, as it complicates the task of determining each weight’s importance and the behavior of the pruning process. As a result, most current state-of-the-art strategies prioritize empirical results through heuristics, often at the expense of understanding the pruning process. 

Given this gap, we introduce Hyperflux, a novel $L_0$ pruning method that shifts the focus from empirical results to understanding the pruning process. Our aim within Hyperflux is to answer the following questions: Why do weights get pruned or regrown? And how does the network behave as we prune it?

Hyperflux is inspired by the principle that \textit{the value of something is not truly known until it is lost}, which has shaped major discoveries in fields such as functional genomics \citep{giaever2002functional, shalem2014genome}, neuroscience \citep{rorden2004using}, and network science \citep{albert2000error}. 

The main idea of our method is to evaluate each weight's importance by first removing it, so we associate each weight with a presence parameter used to mask it. Once the weight is masked, we observe the resulting gradient in its absence, which we term weight \textit{flux}. A global \(L_0\) regularization term called \textit{pressure} pushes all weights towards pruning, aiming to uncover each of their fluxes. Those weights whose flux is greater than the pressure will be regrown, while the rest will remain pruned. This process is continuously applied until the end of training and represents a continuously evolving system which explains pruning, as detailed in section \ref{section:Hyperflux}.

Apart from modeling pruning, Hyperflux introduces three novel elements: (i) evaluating weights while pruned, a deviation from the classical first-order criterion, (ii) an  \(L_0\)-type regularization which leads the weights towards evaluation (pruning), and (iii) a novel scheduler for sparsity targeting.

Our main contributions are:
\begin{itemize}
    \item Hyperflux, an \(L_0\) method which introduces a novel pruning formulation, along with a novel scheduler for targeting desired sparsities. 
    \item Understanding and modeling Hyperflux's pruning process at microscopic and macroscopic levels. 
    \item Competitive results against well-established pruning methods (e.g. CAP, GraNet, AC/DC), across ResNet-50, VGG-19, and DeiT-T/S on CIFAR-10, CIFAR-100, and ImageNet.
\end{itemize}

\section{Related work}
\label{section:related}

Research on neural network pruning has a long history, with some methods going back decades and laying the groundwork for modern approaches. Early approaches, such as \citet{lecun1989optimal} and \citet{lecun1992second}, utilized Hessian-based techniques and Taylor expansions to identify and remove unimportant specific weights, while \citet{skeletonization1988} employed derivatives to remove whole units, an early form of structured pruning. These initial studies demonstrated the feasibility of reducing network complexity without significantly compromising performance. An influential overview \citep{thimm1995evaluating} concluded that magnitude pruning was particularly effective, a paradigm that has since been widely adopted \citep{han2016deepcompression, frankle19lottery, zhou19deconstructing, evci2020rigging, kusupati2020soft, han2015learning, tai2022spartan, peste2021acdc, georgoulakis2023feather}.

\textbf{The existence of highly effective subnetworks} builds upon these foundational studies, with the Lottery Ticket, \citep{frankle19lottery}, being a good example. This work uses magnitude pruning to demonstrate that there exists a mask which, if applied at the start of training, produces a sparse subnetwork capable of matching the performance of the original dense network after training, if the initialization is kept unmodified. Subsequent research has further validated this concept by showing that these subnetworks produced by masks, even without any training, achieve significantly higher accuracy than random chance \citep{zhou19deconstructing}, reaching up to 80\% accuracy on MNIST. Moreover, training these masks instead of the actual weight values can result in performance comparable to the original network \citep{ramanujan2020hidden, zhou19deconstructing}, suggesting that neural network training can occur through mechanisms different from weight updates, including the masking of randomly initialized weights. Other studies have attempted to identify the most trainable subnetworks at initialization. SNIP~\citep{lee2019snip} uses gradient magnitudes as a way to identify trainable weights, while \citet{savarese2020winningcontinuous} employs \(L_{0}\) regularization along with a sigmoid function that gradually transitions into a step function during training, enabling continuous sparsification. These findings indicate that the specific values and even the existence of certain weights may be less critical than previously believed.

\textbf{Differentiable pruning} literature differs from classical heuristics by incorporating learnable parameters that guide pruning decisions through gradient-based optimization during training. For example, \citet{kusupati2020soft} trains a magnitude threshold for each layer to determine which weights are pruned. Other works, such as \citet{cho2024pdp}, use no explicit pruning parameters, instead learning a weight distribution whose shape implicitly determines sparsity. A related class of methods uses \(L_{0}\) regularization \citep{savarese2020winningcontinuous, louizos2018l0}, which encourages sparsity by penalizing the number of non-zero weights directly, typically through per-weight learnable gate parameters. Hyperflux aligns with the \(L_{0}\) differentiable pruning 
paradigm by enabling continuous pruning of weights based on learnable parameters. However, unlike typical such methods, in which the regularization coefficient \(\lambda\) is fixed and sparsity emerges purely from gradient descent, Hyperflux explicitly controls the pruning pressure through a scheduler \textsc{Sched}\((d_e, e)\) that adapts based on 
the current network density and training epoch, allowing explicit external control over the pruning behavior throughout training. A detailed comparison of Hyperflux against representative differentiable mask methods is provided in Appendix~\ref{app:diff_mask}.

\textbf{Pruning based on gradient values} is another prominent approach that assesses weight properties in relation to the loss function and often overlaps with differentiable methods. Works by \citet{lee2019snip} and \citet{dejorge2021progressiveforce} assess the trainability of subnetworks by analyzing initial gradient magnitudes relative to the loss function. \citet{xia2019autoprune} introduces handcrafted gradients that influence training, while \citet{lin2020dynamic} uses gradients during backpropagation to recover pruned weights with high trainability, preserving accuracy. \citet{evci2020rigging} uses gradient and weight magnitudes to determine which weights to prune and to regrow. \citet{liu2021sparse} employs a neuroregeneration scheme, which prunes and regrows the same number of weights, effectively keeping the sparsity constant while growing accuracy. Hyperflux distinguishes itself from all these methods by evaluating the importance of weights \textit{after} the moment of their pruning. Instead of deciding which weights are (un)important based solely on instantaneous gradients or single-stage evaluations, Hyperflux identifies a weight's significance based on the impact its removal has on the network's performance, aggregated across topologies. 

\section{Hyperflux method}
\label{section:Hyperflux}

We associate each weight \(\omega_{i}\) with a learnable parameter \(t_{i}\), which determines whether the weight is present (\(t_i>0\)) or pruned (\(t_i \leq 0\)). We define a weight's importance to be the increase in loss caused by its pruning. We assess the importance of a weight \(\omega_i\) through its flux, the gradient of the loss function with respect to \(t_i\) when \( t_i\leq0 \). The connection between flux and weight importance is detailed in Section \ref{subsec:weight_flux}. The \textit{pressure} term, denoted by \(L_{-\infty}\), will push all \(t\) values towards \(-\infty\), pruning the weights and revealing their fluxes. No manual selection or analysis of gradients is needed, since the interaction between pressure and flux during backpropagation will naturally only keep important weights whose flux is large.

\subsection{Preliminaries}
\label{sec:notation}
Consider a neural network defined as a function \(f: \mathcal{X} \times \mathbb{R}^d \to \mathcal{Y} \), where \(\mathcal{X}\) is the input space, \(\mathcal{Y}\) is the output space, and \(\mathbb{R}^d\) is the space of weights. Given a training set \(\{(x_j, y_j)\}_{j=1}^J\), learning the weights \(\omega\) amounts to minimizing a loss function so that \(f(x_j, \omega)\) aligns with \(y_j\): 
\[
\mathcal{L}(\omega) = \sum_{j=1}^J \ell\bigl(f(x_j, \omega), y_j\bigr).
\]
We define the topology of the neural network as a binary vector \( \mathcal{T} \in \{0, 1\}^d \), where \(\mathcal{T}_i\) represents whether weight \(\omega_i\) is pruned or not. We denote a family of topologies as \(\mathcal{T}^{1 \to K}\), with \(K\) its cardinality and \(\mathcal{T}^k\) a specific topology from the family. Thus, the loss of a network with topology \(\mathcal{T}\) is: 
\[
\mathcal{L}(\omega, \mathcal{T}) = \sum_{j=1}^J \ell\bigl(f(x_j, \omega \odot \mathcal{T}), y_j\bigr),
\]
where \( \odot \) is the Hadamard product. For each weight \(\omega_{i} \), we introduce a learnable presence parameter \(t_{i}\), with \(t \in \mathbb{R}^d\) denoting the vector collecting all \(t_i\). The vector \(t\) is used to generate the topology \(\mathcal{T} \) with \(\mathcal{T}_i = H(t_i)\), where: 
\[
H(t_i) =
\begin{cases} 
1 & \text{if } t_i > 0, \\
0 & \text{if } t_i \leq 0.
\end{cases}
\]
Thus, if \(t_{i} > 0\), then \(\omega_{i}\) is active, otherwise (when \(t_{i} \leq 0\)), \(\omega_i\) is pruned.
We use a global penalty term \(L_{-\infty}\) to push all \(t_i\) values towards \(-\infty\), which we discuss in detail in Section \ref{subsec:weight_flux}. Our goal is to find a topology \(\mathcal{T}^*\) and set of weights \(\omega^*\) such that the following loss is minimized:
\[
\mathcal{J}(\omega, \mathcal{T}) = \mathcal{L}(\omega, \mathcal{T}) + L_{-\infty}(t).
\]

\subsection{Hyperflux Pruning Process}
\label{subsec:weight_flux}
In this section, we gradually build the Hyperflux pruning process, starting from the notions of flux and pressure. We analyze the dynamics of weight pruning and regrowth, define the criteria for pruning decisions, and explore the convergence of sparsity and its relationship with regularization. 

\subsubsection{Flux and Pressure}

We begin by introducing the notion of flux, evaluated on one topology \(\mathcal{T}\), and develop its connection to weight importance. Since the optimal topology \(\mathcal{T}^*\) is initially unknown, any metric measured on some topology \(\mathcal{T}\) might not be relevant for \(\mathcal{T}^*\). For this reason, we then extend flux to \textit{aggregated flux}, a more informative evaluation based on a family of topologies \(\mathcal{T}^{1 \to K}\). 

We start by defining \(\mathcal{G}_i(\omega,\mathcal{T})\), representing the direction in which \(t_i\) needs to change to minimize the loss for topology \(\mathcal{T}\) and weights \(\omega\): 
\begin{equation}
\label{eq:G_one_topology}
\mathcal{G}_i(\omega, \mathcal{T}) = -\frac{\partial \mathcal{L}(\omega, \mathcal{T})}{\partial t_{i}}, {\forall t_i \in \mathbb{R}}.
\end{equation}

To allow computing (\ref{eq:G_one_topology}) despite the non-differentiable step function \(H(t_i)\), we employ a straight-through estimator for the gradient of \(H\) with respect to \(t_i\), which we denote by \(\text{STE}_H\). We choose \(\text{STE}_H(t_i) = 1\); the simplicity of the estimator is essential for deriving the closed-form equation \ref{eq:ineq_flux}, which forms the mathematical basis of our method. This formulation allows for further analytical study and reveals a direct theoretical link to first-order based pruning (detailed in Appendix~\ref{appendixsecsub:flow_and_hessian}).

To fully understand the implications of \(\mathcal{G}_i\) on updating \(t_i\), we study the gradients composing it. We define \(\theta_i = \omega_i \cdot H(t_i)\), and refer to \(\theta_i\) as effective weight. By rewriting \(\mathcal{G}_i\), we get: 
\[
\mathcal{G}_i(\omega, \mathcal{T}) 
=
\underbrace{
-\frac{\partial \mathcal{L}(\omega, \mathcal{T})}{\partial \theta_{i}} 
}_{=:\mathcal{A}_i}
\cdot 
\frac{\partial \theta_i}{\partial t_{i}}
=
\mathcal{A}_i \cdot \omega_i \cdot \text{STE}_H(t_i)
=
\mathcal{A}_i \cdot \omega_i
.
\]
\(\mathcal{A}_i\) represents the direction in which the effective weight \(\theta_i\) should change to minimize the loss. If \(\mathcal{A}_i\) has the same sign as the weight \(\omega_i\), then \(t_i\) will increase, reinforcing presence. Otherwise, if they have different signs, \(t_i\) will decrease towards pruning. This behavior takes two meanings depending on whether \(t_i \leq 0\) or \(t_i > 0\), which we analyze below. For this purpose, we define \(\mathcal{W}_i=-\frac{\partial \mathcal{L}}{\partial \omega_i}\), the direction in which \(\omega_i\) should change to reduce the loss.

For \(t_i>0\), \(\mathcal{W}_i =\mathcal{A}_i \cdot H(t_i) = \mathcal{A}_i \). Therefore, \( \mathcal{G}_i(\omega, \mathcal{T}) \) can be rewritten as \( \mathcal{W}_i \cdot \omega_i \), meaning that \(t_i\) increases when \(\mathcal{W}_i\) and \(\omega_i\) have the same sign and decreases otherwise. Note that \(\mathcal{W}_i\) and \(\omega_i\) having the same sign also means that \(|\omega_i|\) increases, while opposite signs imply that \(|\omega_i|\) decreases. Therefore, \(t_i\) follows the direction of change in \(|\omega_i|\) if the loss's gradient acted on it alone.

\begin{figure}[htbp]
    \centering
    \begin{subfigure}[b]{0.235\linewidth}
        \includegraphics[width=\linewidth]{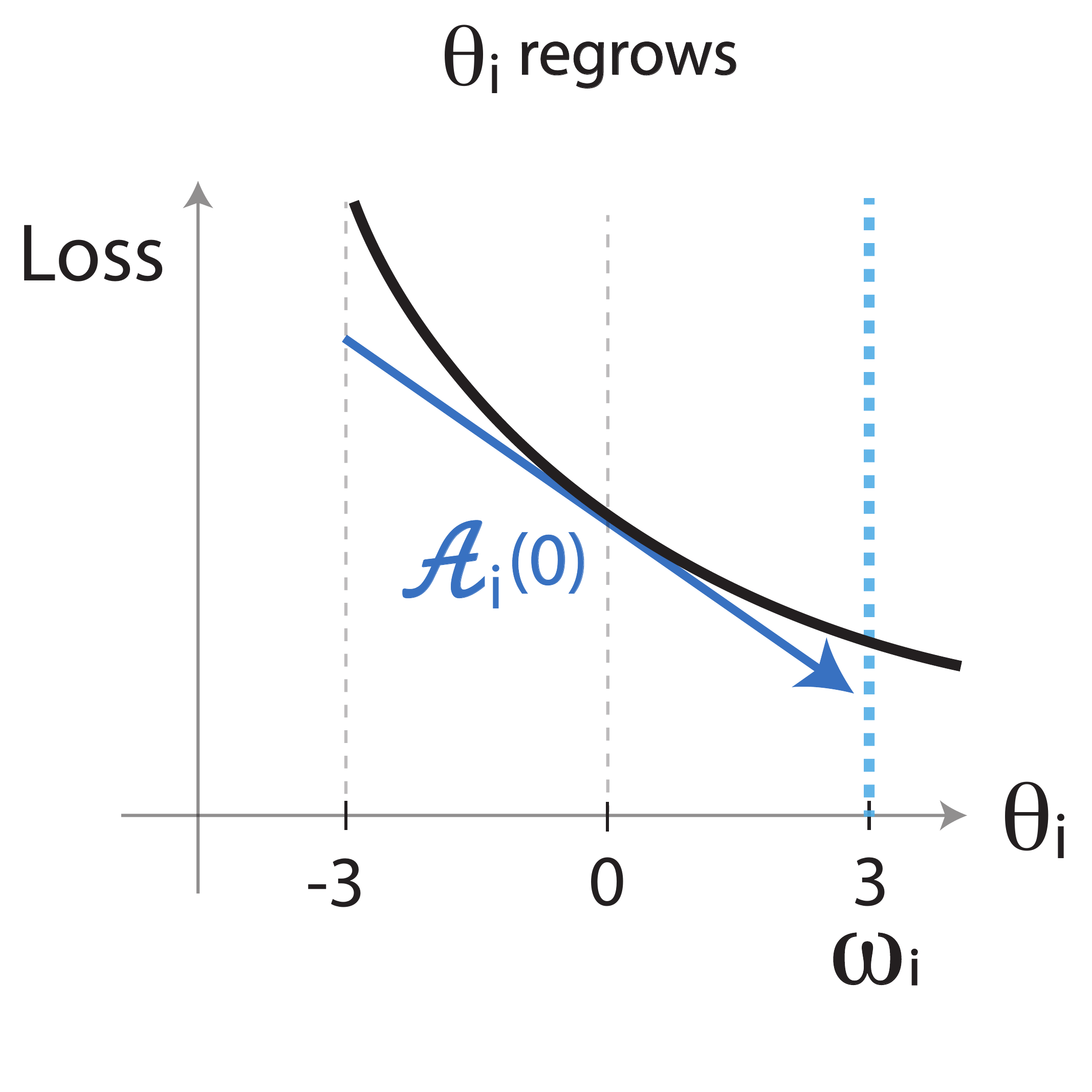}
        \caption{}
        \label{fig:sub1_1}
    \end{subfigure}
    \hfill 
    \begin{subfigure}[b]{0.235\linewidth}
        \includegraphics[width=\linewidth]{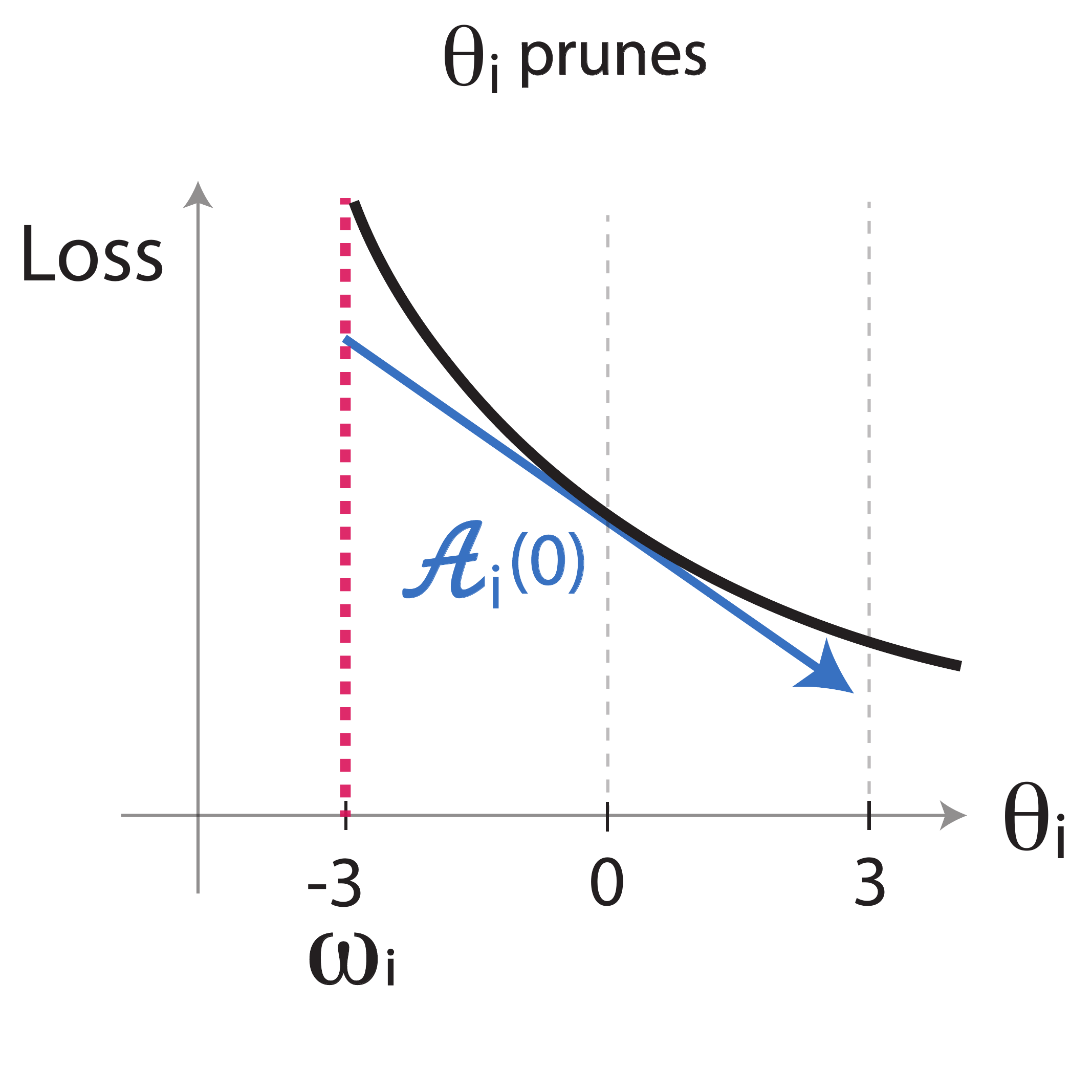}
        \caption{}
        \label{fig:sub2_1}
    \end{subfigure}
    \hfill
    \begin{subfigure}[b]{0.235\linewidth}
        \includegraphics[width=\linewidth]{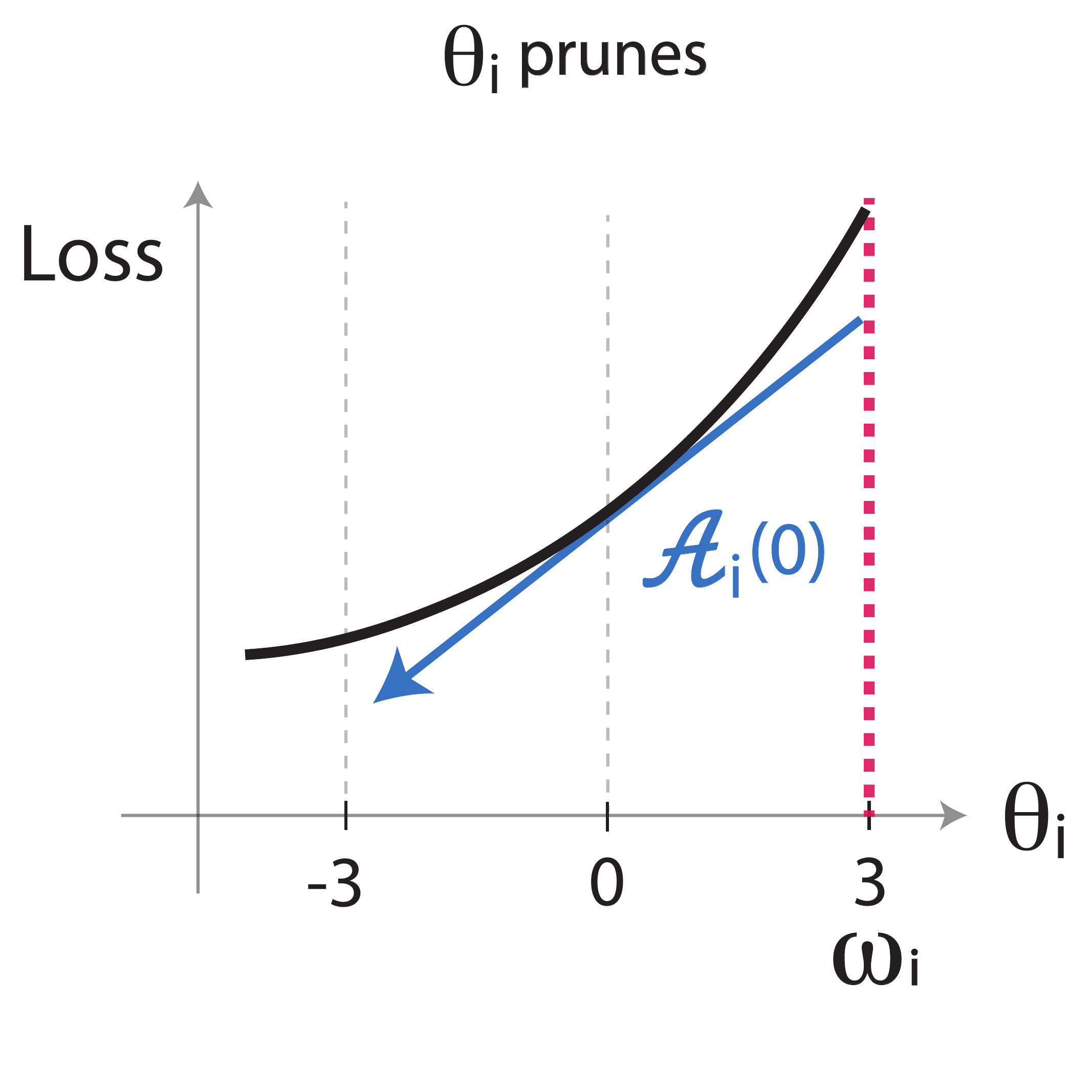}
        \caption{}
        \label{fig:sub3_1}
    \end{subfigure}
    \hfill
    \begin{subfigure}[b]{0.235\linewidth}
        \includegraphics[width=\linewidth]{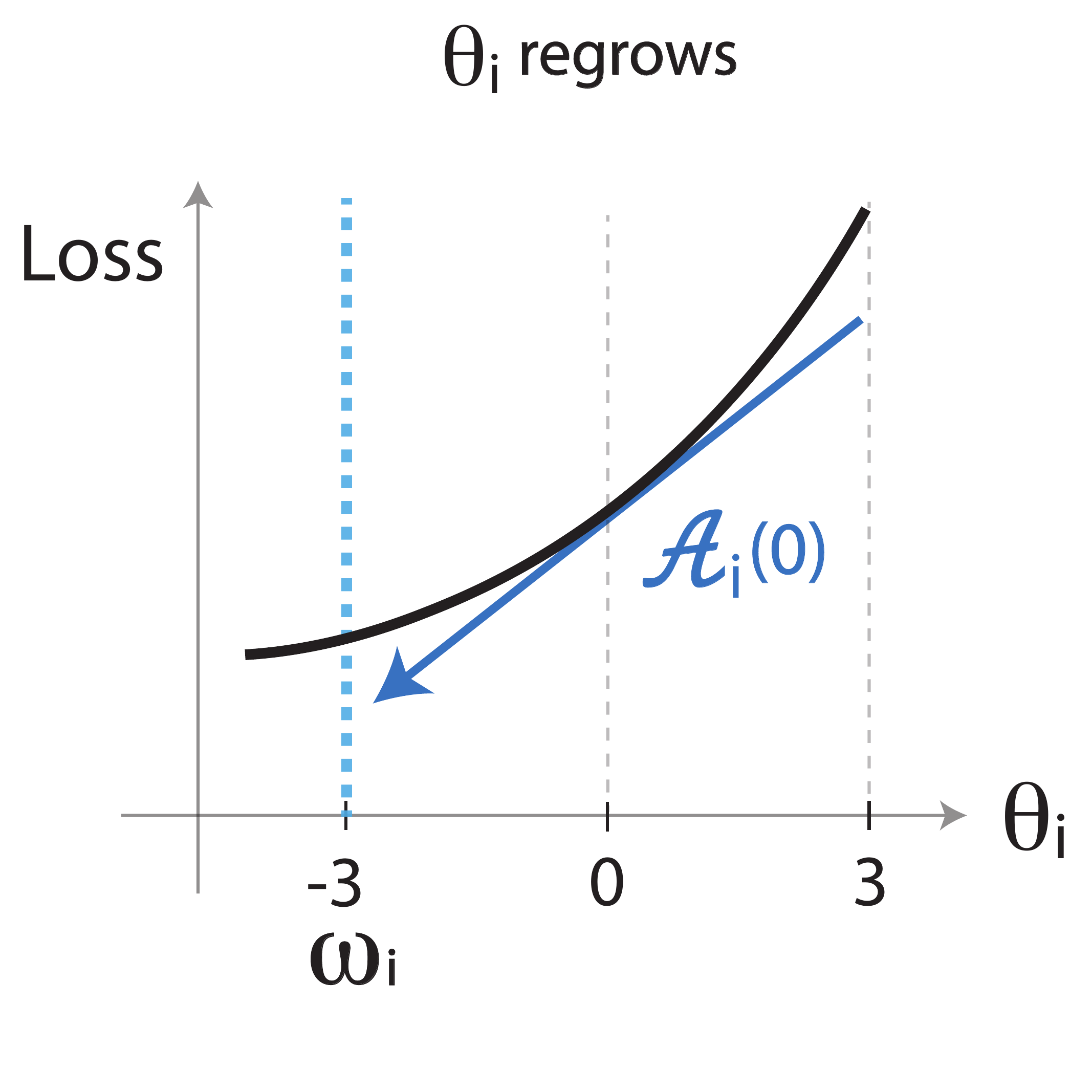}
        \caption{}
        \label{fig:sub4_1}
    \end{subfigure}
    \caption{Scenarios for \(\theta_i\) when \(H(t_i)=0\). If \(\mathcal{A}_i\) points towards \(\omega_i\) the flux \(\mathcal{G}^-_i\) regrows the weight, as in (a) and (d). Otherwise, it keeps the weight pruned, as in (b) and (c). Numerical values are only illustrative. }
    \label{fig:four-diagrams}
\end{figure}

To assess the importance of \(\theta_i=\omega_i\), the method allows \(t_i\leq0\), causing \(\theta_i=0\), and checks whether, as a result, \(\mathcal{A}_i\) points towards \(\omega_i\), i.e., whether \(\text{sign}(\mathcal{A}_i) = \text{sign}(\omega_i)\). If this is true, moving \(\theta_i\) from $0$ towards $\omega_i$ would reduce the new loss (obtained after $\theta_i$ became $0$) and consequently, \(\mathcal{G}_i\) increases \(t_i\) until regrowth, \(\theta_i=\omega_i\). In this way, Hyperflux implements the key insight that \textit{one never knows the value of something (\(\theta_i\)) until one loses it (sets it to 0).} Otherwise, if \(\text{sign}(\mathcal{A}_i) \neq \text{sign}(\omega_i)\), \(t_i\) decreases, keeping the weight pruned, \(\theta_i=0\). All four combinations of signs are presented in Fig.~\ref{fig:four-diagrams}.  For this \(t_i \leq 0\) setting, \(\mathcal{G}_i\) takes the meaning of \textit{flux}, and its relation to weight importance is further discussed in Appendix \ref{appendixsecsub:weightimportance_largerflux}.



Given the fact that \(\mathcal{G}_i(\omega, \mathcal{T})\) takes two different meanings, we introduce two different notations:
\begin{equation}
\label{eq:G_split}
\mathcal{G}_i(\omega, \mathcal{T}) =:
\begin{cases} 
\mathcal{G}_i^-(\omega, \mathcal{T}), &t_i \leq 0, \\
\mathcal{G}_i^+(\omega, \mathcal{T}), &  t_i > 0.
\end{cases}
\end{equation}
\(\mathcal{G}_i^-(\omega, \mathcal{T})\) refers to flux, whereas \(\mathcal{G}_i^+(\omega, \mathcal{T})\) is the direction of change of \(|\omega_i|\). While the instantaneous flux shares mathematical similarities with first-order saliency metrics such as those of~\citet{lee2019snip} and ~\citet{molchanov2017pruning}, the paradigm differs by being applied to pruned weights instead of active ones. This is a key part of Hyperflux's novelty and detailed more in Appendix~\ref{app:flux_saliency}.

To drive \(t\) values towards \(-\infty\), we employ an ``\(L_{-\infty}\)'' loss called \textit{pressure}, formulated as:
\begin{equation}
\label{eq:pressure_initial}
L_{-\infty}(t) = \frac{1}{d} \cdot \gamma \cdot \sum_{i=1}^{d} \, t_{i},
\end{equation}
where \(\gamma\) is a scalar used to control sparsity and \(d\) the total number of weights in the network. Any reference about an increase, decrease, value, or scheduler of pressure will refer to \(\gamma\). The pressure term yields a constant gradient \(\frac{\gamma}{d}\) with respect to each \(t_i\) parameter, independent of their current value. Unlike other methods which choose to apply custom pruning schedules \citep{evci2020rigging, frankle19lottery, liu2021sparse}, we let the interaction between \(L_{-\infty}(t)\) and \(\mathcal{G}^+_i(\omega, \mathcal{T})\) decide which weights get pruned and therefore have their fluxes uncovered. We compare to \citep{molchanov2017pruning, lee2019snip}, and emphasise the difference from Taylor-based methods in Appendix \ref{app:flux_saliency}.

\subsubsection{Aggregated flux}
\label{subsubsec:aggregated_flux}

\begin{wrapfigure}{r}{0.35\textwidth}
    \centering
    \includegraphics[width=\linewidth]{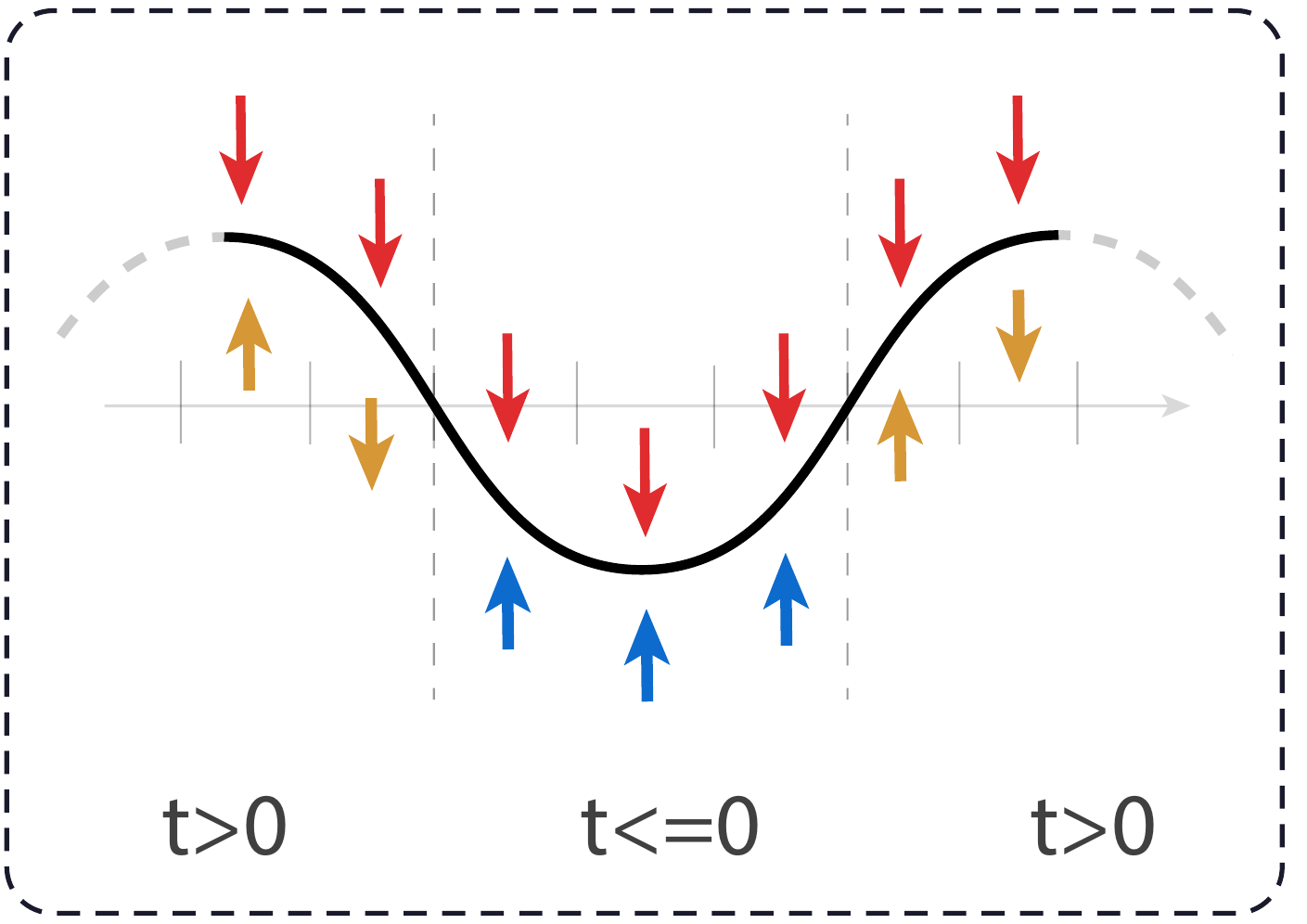}
    \caption{Depiction of gradients (as arrows) influencing $t_i$, red, yellow and blue denote pressure, $\mathcal{G}_i^+$ and $\mathcal{G}_i^-$ respectively.}
    \label{fig:t_lifecycle}
\end{wrapfigure}

We let \(\mathcal{G}_i(\omega, \mathcal{T})\) and the gradient of \(L_{-\infty}(t)\) interact during backpropagation without direct intervention. As a result, a family of topologies \( \mathcal{T}^{1 \to K} \) \textit{emerges implicitly} during training by concurrent pruning (determined by \(L_{-\infty}\)) and regrowth (determined by \(\mathcal{G}^-_i\) increasing \(t_i\)). Furthermore, a \(t_i \leq 0\) may be increased for several iterations until it reaches \(t_i>0\), being evaluated at each iteration over a potentially different topology \(\mathcal{T}^k \in  \mathcal{T}^{1 \to K}\). This behavior is desirable, given that evaluating flux on a single topology provides a limited estimate of importance. To get a better picture of the underlying interactions, we begin by extending equation (\ref{eq:G_split}) to a family of topologies: 
\begin{equation}
  \label{eq:aggregated_flux}
  \mathcal{G}_i^{-/+}\!\Bigl(\omega, \mathcal{T}^{1 \to K}\Bigr)
  = \frac{1}{K}\,\sum_{k=1}^K
    \mathcal{G}_i^{-/+}\!\Bigl(\omega, \mathcal{T}^k\Bigr).
\end{equation}

This leads to an \textit{aggregated flux} \(\mathcal{G}_i^-(\omega, \mathcal{T}^{1 \to K})\) and an average tendency of change in weight magnitude \(\mathcal{G}_i^+(\omega, \mathcal{T}^{1 \to K})\) respectively. In Hyperflux, the updates over \(H\) iterations can be written as: 

\begin{equation}
\label{eq:gradient_t_extended_sum}
\sum_{h=1}^{H} \frac{\partial (\mathcal{L}(\mathcal{T}^{h}, \omega) + L_{-\infty}(t))}{\partial t_{i}} = 
\sum_{h=1}^{H} (-\mathcal{G}_i(\omega,\mathcal{T}^{h}) + \frac{\gamma}{d}),
\end{equation}

where \(\mathcal{T}^h\) is the topology at iteration \(h\). We examine the ``life cycle'' of a presence parameter \(t_i\) over the $H$ training iterations. In Figure \ref{fig:t_lifecycle} we show how the gradients of \(t_i\), represented by arrows, interact. During these $H$ steps, \(t_i\) alternates between active phases during which it follows the tendency of \(|\omega_i|\), and pruned phases during which flux accumulates. We refer to the transition from a pruned phase back to a present phase as \emph{implicit regrowth}. To illustrate the interactions between flux and pressure in our method, consider a pruned phase beginning at iteration $P_s$ and ending at iteration $P_f$ ($1 < P_s < P_f \le H$). If $P_f$ marks the final step of that pruned phase, the total change in $t_i$ over $[P_s,P_f]$ is positive, which gives:
\begin{equation}
\label{eq:ineq_flux}
    \sum_{h=P_s}^{P_f} (\mathcal{G}_i(\omega,\mathcal{T}^h) - \frac{\gamma}{d}) > 0 
\end{equation}

This condition holds if and only if: 
\begin{equation}
\label{eq:ineq_pressure_condition}
    (P_f-P_s) \cdot \Bigl[\mathcal{G}_i^{-}\!\Bigl(\omega, \mathcal{T}^{P_s \to P_f}\Bigr) - \frac{\gamma}{d}\Bigr] > 0.  
\end{equation}

Thus, a weight will be regrown if the aggregated flux is greater than the pressure. Conversely, after an active interval, the weight becomes pruned, if
\((P_f-P_s) \cdot\Bigl[\mathcal{G}_i^{+}\!\Bigl(\omega, \mathcal{T}^{P_s \to P_f}\Bigr) - \frac{\gamma}{d}\Bigr] < 0\). The practical limits and interactions with weight decay are discussed in Appendix \ref{app:PruningProcessDisc}. For brevity, we write the sequel \(\hat{\mathcal{G}}_i^{\pm} := \mathcal{G}_i^{\pm}(\omega, \mathcal{T}^{1 \to K})\) for the aggregated quantities.

A weight $\omega_i$ is pruned if its gradient either (i) drives it toward zero ($\widehat{\mathcal{G}}_i^+ < 0$) or (ii) fails to pull away from 0 sufficiently ($0 < \widehat{\mathcal{G}}_i^+ < \frac{\gamma}{d}$). Conversely, a weight resists pruning when its gradient pulls it away from zero strongly enough ($\widehat{\mathcal{G}}_i^+ > \frac{\gamma}{d}$). Pruning does not measure importance, but selects plausible candidates \textit{for} evaluating importance, see again Appendix \ref{app:PruningProcessDisc}. Importance is evaluated after pruning, as discussed next. 


A weight gets regrown, due to being deemed important, when the pressure is smaller than the increase in loss caused by removal, i.e., whenever $\widehat{\mathcal{G}}_i^- > \frac{\gamma}{d}$, and the weight encoded useful information. Conversely a weight with small or negative $\widehat{\mathcal{G}}_i^-$ was either unimportant or actively harmful, and stays pruned. The sign-based mechanism underlying this behavior is illustrated in Fig.~\ref{fig:four-diagrams}. The empirical behavior of weights is seen in Appendix \ref{appendixsecsub:implicit-regrowth}, supporting our predictions. 

 These findings provide the answer to the first question from the introduction: Why do weights get pruned or regrown? In our experiments, since our method relies on weights that already encode meaningful information, we begin pruning by initializing the network with \textit{pretrained weights}.

\subsubsection{Minimum Flux and Convergence}
\label{subsec:experimental_properties}

Following from the insights about flux and pressure described so far, we postulate a series of empirical properties that naturally emerge. We experimentally validate each of the properties.

\begin{figure}[htbp]
  \centering
  \includegraphics[width=0.4\textwidth]{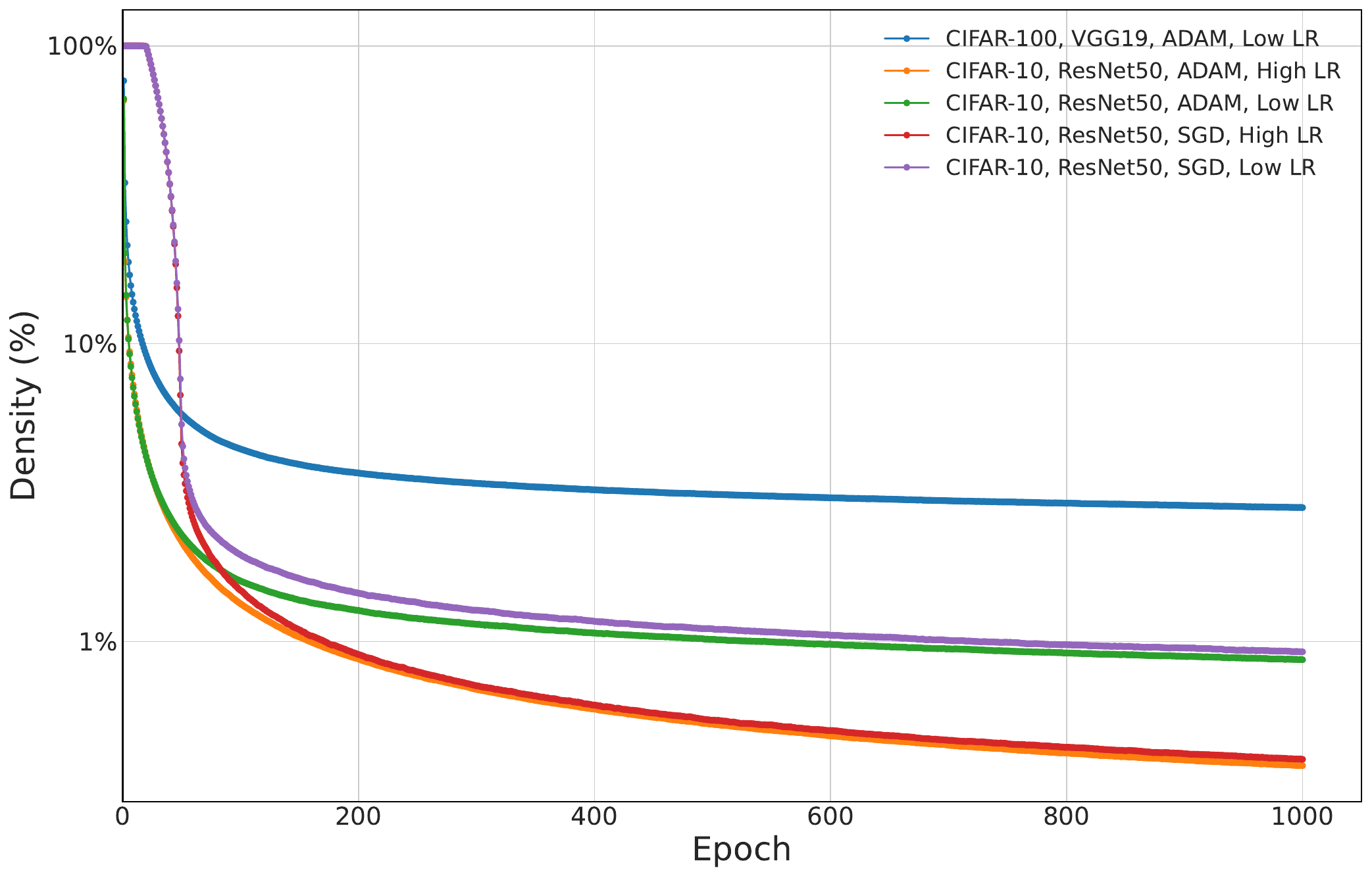}
  \caption{Convergence for fixed \(\gamma = 2\).}
  \label{fig:sparsity_fixed_gamma}
\end{figure}

\textbf{Property 1: Density Convergence for a Fixed \(\gamma\)}. As density decreases, fewer weights are used to represent the same information contained within the dataset. As long as the accuracy remains constant, the importance -- and therefore flux -- of the remaining weights should increase. Once the minimum flux among all weights, \(\mathcal{W}_f\), matches the pressure, we expect density to converge to an equilibrium. Therefore, we ask the following question: \textit{Given a fixed \(\gamma\), will the network converge to a final density \(\mathcal{D}\)?}
In Figure~\ref{fig:sparsity_fixed_gamma}, we test this by running ResNet-50 and VGG-19 on CIFAR-10 and CIFAR-100 respectively. We preload weights for each network and allow them to train for 1000 epochs with a constant pressure \(\gamma\).

To better understand how convergence is affected by optimizers and learning, we test two different optimizers for \(t\) values, SGD and Adam, along with a high and low learning rate, differing by two orders of magnitude. We observe that networks with low learning rates tend to converge faster and smoother compared to high learning rates. Further ablation studies are found in Appendix~\ref{appendixsec:ablation_studies}.

\begin{figure}[!htbp]
  \centering
  \begin{subfigure}[b]{0.4\textwidth}
    \centering
    \includegraphics[width=\linewidth]{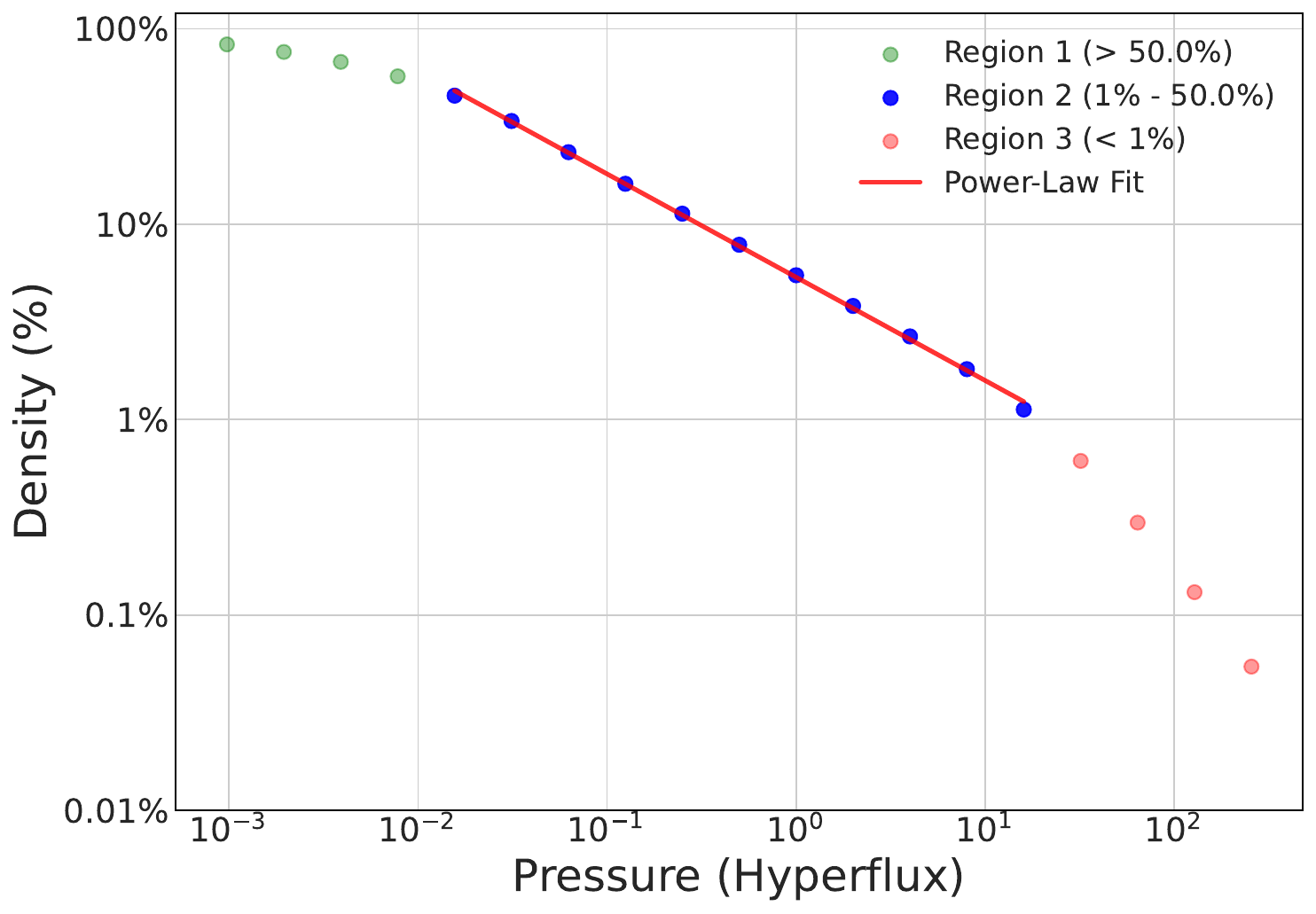}
    \caption{VGG-19, CIFAR-100}
    \label{fig:vgg19c100nplhHyperflux}
  \end{subfigure}
  \hspace{0.05\textwidth} 
  \begin{subfigure}[b]{0.4\textwidth}
    \centering
    \includegraphics[width=\linewidth]{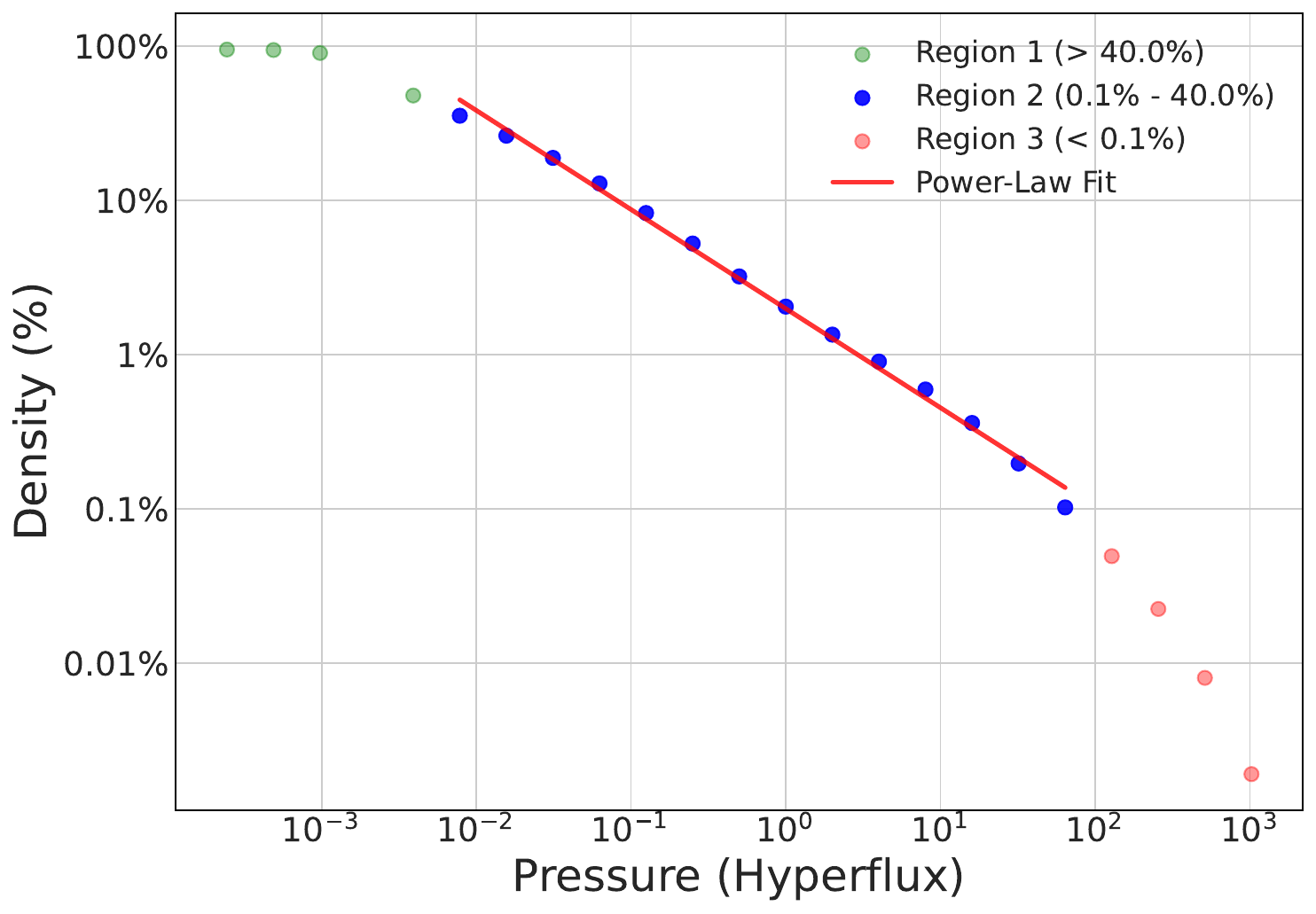}
    \caption{ResNet-50, CIFAR-10}
    \label{fig:vgg19c100nplhIMP}
  \end{subfigure}
  \caption{The relationship between \(\gamma\) (minimum flux) and final density for ResNet-50, CIFAR-10 and VGG-19 CIFAR-100. The 3 regions we discussed are highlighted in the figure.}
  \label{fig:gamma_final_sparsity}
\end{figure}

\textbf{Property 2: Relationship Between Minimum Weight Flux and Density}. Property 1 says that all networks converge to a fixed density for a fixed \(\gamma\), when \(\mathcal{W}_f = \gamma\), and do not end up diverging or oscillating. Therefore, we ask: \textit{Is there a predictable relationship between minimum flux among weights and network density?} For each experiment, we train a network with constant \(\gamma\) and allow it to converge. The learning rate is kept constant, since it influences the final density (see Fig \ref{fig:sparsity_fixed_gamma}). Figure~\ref{fig:gamma_final_sparsity} illustrates different convergence points for several \(\gamma\) values for ResNet-50 on CIFAR-10, and VGG-19 on CIFAR-100. 

We observe three regions taking shape in our experiments. In region 1, the pressure is too small for any meaningful weights to be pruned and information compression to occur. In region 2, we observe constant accuracy over a large density range, where meaningful information compression occurs, leading to a clear power-law relationship between pressure and density.
\begin{equation}
\label{eq:relationship-pressure-sparsity}
\ln(s) \;=\; \ln(c) \;-\; \alpha_0\,\ln(\gamma)
\end{equation}
where the constants $c$ and \(\alpha_0\) depend on dataset and network architecture. In region 3, accuracy starts collapsing along with the remaining weights, leading eventually to 0\% density. Power-laws are frequently found in deep learning, as proven by neural scaling laws \citep{hestness2017scaling, kaplan2020scaling}, loss behavior during magnitude pruning \citep{rosenfeld2021predictability} and weight distributions \citep{martin2021implicit}. 

Therefore, we answer the second question in the introduction: How does the network behave as we prune it? Properties 1 and 2 together explain the macroscopic behavior of Hyperflux. Property 1 establishes that, for a fixed pressure \(\gamma\), the network converges to a stable equilibrium density at which the minimum aggregated flux among surviving weights matches the pressure. Property 2 then characterizes the dependence of this equilibrium on \(\gamma\), revealing a clean power-law between pressure and final density, which shows that sparsity and the pruning criterion co-evolve in a predictable and reproducible manner determined by \(\gamma\). 

This predictability makes sparsity targeting feasible, motivating the pressure scheduler introduced next, in which power-law updates to \(\gamma\) translate directly into controlled changes in density.


\subsection{Pressure Scheduler}
\label{subsec:pressure_scheduler}

Achieving a specific target sparsity in $L_0$ regularization has historically posed significant challenges. Previous approaches often rely on expensive grid searches over the parameter space or require extended training schedules to allow smooth network convergence \citep{louizos2018l0, savarese2020winningcontinuous}. Lagrangian formulations \citep{gallego2022controlled} attempt to adjust the regularization during training, but they frequently produce unstable and oscillatory behavior. Recent work by \citet{sohrabi2024pi} addressed these issues by employing a Proportional-Integral (PI) controller, borrowed from control engineering, to adjust regularization. 

We propose a scheduler alternative that, similarly to PI controllers, uses network feedback for adjustment. We define the pressure as $\gamma_e = (p_e)^\alpha$, where base $p_e$ is a scalar updated according to Algorithm~\ref{alg:sparsity_scheduler}. To maintain stability against suboptimal hyperparameter choices for the step size $u$ or exponent $\alpha$, we incorporate inertia terms $p_+$ and $p_-$, similar to the integral effect in PI controllers. However, unlike standard PI controllers which (i) use linear increments and (ii) always minimize a regulation error, our scheduler (i) applies updates to $\gamma$ in power-law increments ($\gamma_e = (p_e)^\alpha$, based on the mathematical properties outlined in Section \ref{subsec:experimental_properties}) and (ii) has a customizable policy $\Pi$ covering various needs (e.g., prioritizing a specific pruning trajectory or precision in the final target sparsity). Concrete policy examples are provided in Appendix \ref{appendixsec:schedulers-implementation}.

The scheduler still requires an initial parameter search. However, once found, the parameters work well for an entire class of similar training tasks (e.g., vision) in our experiments.


\begin{algorithm}[htbp]
\caption{Hyperflux Pruning Algorithm}
\label{alg:Hyperflux_simplified} 
\small
\begin{algorithmic}[1]
\State \textbf{Input:} Pretrained weights $\omega^{\mathrm{init}}$, pruning epochs $E_p$, stabilization epochs $E_s$ (leading to total epochs $E_t = E_p + E_s$), pressure scheduler {\sc Sched}($d_e,e$).
\State \textbf{Output:} Weights $\omega^*$, final topology $\mathcal{T}^*$.

\State \textbf{Initialize:}
\State \quad Weights $\omega \leftarrow \omega^{\mathrm{init}}$ .
\State \quad Presence parameters $t_i \gets$ positive values, $\forall i \in \{1,2,...,d\}$.
\State \quad Topology $\mathcal{T}_i \leftarrow 1$, $\forall i \in \{1,2,...,d\}$.
\For{epoch $e = 1$ to $E_t$}
    \State Calculate total loss $\mathcal{J}(\omega, \mathcal{T}) = \mathcal{L}(\omega, \mathcal{T}) + L_{-\infty}(t)$.
    \State $\omega \gets \omega - \eta_\omega \nabla_\omega \mathcal{L}$.
    \State $t \gets t - \eta_t \nabla_t \mathcal{J}$.

    \If{$e \leq E_{p}$}
        \State $\gamma \gets \text{{\sc Sched}}(\text{current density }d_{e}, e)$ 
    \Else
        \State $\eta_t \gets 0.9 \cdot \eta_t$
        \State $\gamma \gets 0$ 
    \EndIf
\EndFor
\end{algorithmic}
\end{algorithm}

\textbf{Stabilization Stage:} One side effect of Hyperflux is the noise created by pruning and reactivation of weights, which, while helpful for pruning, is harmful for convergence. For this reason, to allow the weights and network topology to converge, we introduce a stabilization stage meant to boost accuracy. Specifically, we set the pressure to zero to encourage regrowth while simultaneously decaying the learning rate \(\eta_t\) to prevent excessive reactivation. The full Hyperflux Pruning algorithm can be seen in Algorithm \ref{alg:Hyperflux_simplified}.

\begin{algorithm}[htbp]
\caption{Pressure scheduler - {\sc Sched}($d_e, e$)}
\label{alg:sparsity_scheduler}
\small
\begin{algorithmic}[1]

\State \textbf{Input: } Current density $d_e$ and epoch $e$
\State \textbf{Requires:} Pruning epochs $E_p$, desired final density $\mathcal{D}$, pressure policy $\Pi$, step $u$, exponent $\alpha$.
\State \textbf{Internals:} Positive and negative inertia $p_+$, $p_-$, base scalar $p_e$ required for epoch 1, $p_0$ (all initialized to 0).
\Statex $\triangleright$ \emph{Runs after each epoch}
\If{$\Pi(E_p, \mathcal{D}, d_e,e)$}
    \State $p_e \gets p_{e-1} + u + p_+$
    \State $p_+ \gets p_+ + \frac{u}{4}$
    \State $p_- \gets 0$
\Else 
    \State $p_e \gets p_{e-1} - u - p_-$
    \State $p_- \gets p_- + \frac{u}{4}$
    \State $p_+ \gets 0$
\EndIf
\State $p_e \gets \max(p_e, 0)$
\State \textbf{Return}: pressure $\gamma_e = (p_e)^{\alpha}$
            
\end{algorithmic}
\end{algorithm}

\section{Hyperflux Performance}
\label{experimental}

We now show that Hyperflux, in addition to giving us a model for understanding the pruning process, also has strong empirical performance. To validate \textit{Hyperflux}, we conduct comprehensive pruning experiments on a diverse set of architectures and datasets: ResNet‑50 and VGG‑19 on CIFAR‑10/100, ResNet‑50 and DeiT-Tiny/Small on ImageNet‑1K. We pit Hyperflux against state-of-the-art pruning approaches such as \citet{liu2021sparse,tai2022spartan,kuznedelev2023cap, peste2021acdc}. To ensure a fair comparison, we run all other methods ourselves, initializing them with the pretrained weights used in Hyperflux, while maintaining the same training budget and augmentations. We test several training setups for each method and report the best results, to ensure no unfair degradation occurs due to suboptimal hyperparameters.

Additionally, to better position Hyperflux within the broader literature, we choose to include one-shot methods \citep{lee2019snip, grasp2020chaoqi, synflow} commonly used as benchmarks in other works, even though our post-training setup is not applicable to them. These benchmarks will be marked with \(*\). 

None of our comparison methods incorporate learnable masks as Hyperflux does. Although we identified some mask-based methods \citep{savarese2020winningcontinuous, louizos2018l0, zhang2022optg}, the missing code and major differences in benchmarks prevent a direct comparison to our work. Each configuration but ResNet-50 and DeiT on ImageNet is run three times, and we report the results as mean \(\pm\) standard deviation; all experiments are run on three NVIDIA GeForce RTX 4090 GPUs. Full details on the training recipe are in~Appendix~\ref{appendixsec:training_setup}. 
\subsection{CIFAR-10 / 100}

\begin{table*}[ht]
\centering
\caption{Comparison on CIFAR-10 and CIFAR-100 datasets at different pruning ratios (90.0\%, 95.0\%, 98.0\%). Bold values represent the best mean performance for each setting.}
\label{table:comparison_all}
\resizebox{1.0\textwidth}{!}{
\begin{tabular}{lccc ccc}
\toprule
\textbf{Dataset} & \multicolumn{3}{c}{CIFAR-10} & \multicolumn{3}{c}{CIFAR-100} \\
\cmidrule(lr){2-4} \cmidrule(lr){5-7}
Pruning ratio & 90.0\% & 95.0\% & 98.0\% & 90.0\% & 95.0\% & 98.0\% \\
\midrule
\textbf{VGG-19}~(Dense) &  & 93.85 $\pm$ 0.06 &  &  & 73.44 $\pm$ 0.09 &   \\
\midrule
SNIP$^*$ & 93.63 & 93.43 & 92.05 &72.84	& 71.83 & 58.46 \\
GraSP$^*$ & 93.30 & 93.04  & 92.19 & 71.95 & 71.23 & 68.90 \\
Synflow$^*$ & 93.35 & 93.45 & 92.24 & 71.77 & 71.72 & 70.94 \\
GMP & 93.82 $\pm$ 0.15 & 93.84 $\pm$ 0.14 & 92.34 $\pm$ 0.13 & 73.57 $\pm$ 0.20 & 73.39 $\pm$ 0.11 & 72.78 $\pm$ 0.07 \\
RigL & 93.60$\pm$0.15 & 93.17$\pm$0.09  & 92.39 $\pm$ 0.04 & 73.03$\pm$0.14  & 72.68$\pm$0.22  & 70.02 $\pm$ 0.7 \\

GraNet ($s_i = 0$) & 93.87 $\pm$ 0.05 & 93.84 $\pm$ 0.16 & 93.87 $\pm$ 0.11 & 74.08 $\pm$ 0.10 & 73.86 $\pm$ 0.04 & \textbf{73.00 $\pm$ 0.18} \\

Hyperflux (ours) & \textbf{94.05 $\pm$ 0.17} & \textbf{94.15 $\pm$ 0.14} & \textbf{93.95 $\pm$ 0.18} & \textbf{74.37 $\pm$ 0.21} & \textbf{74.18 $\pm$ 0.15} & 72.9 $\pm$ 0.05 \\

\midrule
\textbf{ResNet-50}~(Dense) &  & 94.72 $\pm$ 0.05 &  &  & 78.32 $\pm$ 0.08 &  \\
\midrule
SNIP$^*$ & 92.65 & 90.86 & 87.21 & 73.14 & 69.25 & 58.43 \\
GraSP$^*$ & 92.47 & 91.32  & 88.77 & 73.28 & 70.29  & 62.12\\
Synflow$^*$ & 92.49& 91.22 & 88.82 & 73.37 & 70.37 & 62.17 \\

RigL & 94.02$\pm$0.33 & 93.76$\pm$0.23  & 92.93 $\pm$ 0.1 & 78.04$\pm$0.19  & 77.39$\pm$0.21 & 75.61 $\pm$ 0.11 \\

GMP & 94.81 $\pm$ 0.05 & 94.89 $\pm$ 0.1 & 94.52 $\pm$ 0.12 & 78.39 $\pm$ 0.18 & 78.38 $\pm$ 0.43 & 77.16 $\pm$ 0.25 \\

GraNet ($s_i = 0$) & 94.69 $\pm$ 0.08 & 94.44 $\pm$ 0.01 & 94.34 $\pm$ 0.17 & 79.09 $\pm$ 0.23 & 78.71 $\pm$ 0.16 & \textbf{78.01 $\pm$ 0.20} \\

Hyperflux (ours) & \textbf{95.41 $\pm$ 0.12} & \textbf{95.15 $\pm$ 0.11} & \textbf{95.26 $\pm$ 0.13} & \textbf{79.58 $\pm$ 0.18} & \textbf{79.23 $\pm$ 0.16} & 77.7 $\pm$ 0.08 \\

\bottomrule
\end{tabular}}
\end{table*}

We evaluate the performance of \textit{Hyperflux} on CIFAR-10 and CIFAR-100 using ResNet-50 and VGG-19 architectures. Results are presented in Table~\ref{table:comparison_all}. On CIFAR-10, \textit{Hyperflux} outperforms the baseline at 90\%, 95\%, and 98\% sparsity for both VGG-19 and ResNet-50, with accuracy gains under 1\% over the next best. Specifically, for VGG-19, it beats $\mathrm{GraNet}$ by $0.18\%$ and $\mathrm{GMP}$ by $0.23\%$ at 90\% sparsity (rising to $1.61\%$ over $\mathrm{GMP}$ at 98\%), while on ResNet-50 it maintains a $0.7\%$ lead over $\mathrm{GraNet}$ across all levels. We also analyze ResNet-50’s layer-wise sparsity at extreme rates (99.74\%, 99.01\%, 98.13\%) and illustrate weight distribution changes in Appendix~\ref{appendixsecsub:layerwise_sparsity_histo}.

On CIFAR-100, \textit{Hyperflux} leads in 4 of 6 benchmarks, being behind $\mathrm{GraNet}$ by only $0.1\%$ and $0.3\%$ in the other two. Notably, $\mathrm{GraNet}$ gains nearly $2\%$ on ResNet-50 when initialized with our pretrained weights. Conversely, $\mathrm{RigL}$ gains $1.5\%$ points of accuracy on ResNet-50 for CIFAR-100, yet experiences drops of up to $0.3\%$ on ResNet-50 for CIFAR-10. On the remaining two benchmarks, its gains are only moderate. At 90\% and 95\% sparsity, \textit{Hyperflux} outperforms all methods, including $\mathrm{GraNet}$, by $0.5\%$. Furthermore, GMP finds itself at a difference of \(0.2\%\) at \(98\%\) sparsity on VGG-19, increasing to \(1.2\%\) points of accuracy at \(90\%\) sparsity, while RigL is behind by \(2.9\%\) at \(98\%\) and \(1.3\%\) at \(90\%\) sparsity.

\subsection{ImageNet-2012}

\begin{wraptable}[23]{r}{0.53\textwidth}
\centering
\small
\setlength{\tabcolsep}{3pt}
\renewcommand{\arraystretch}{0.9}
\caption{ResNet-50 top-1 accuracy, parameter count, sparsity, and compute cost on ImageNet-2012. We denote by $s$ the sparsity, and by $F_{\mathrm{train}}$ and $F_{\mathrm{test}}$ the compute cost (FLOPs) required for training and testing, respectively.}
\label{table:resnet50_imagenet_compute}
\begin{tabular}{@{}lcccccc@{}}
\toprule
Method & Top-1(\%) & Params & s(\%) & F$_{\mathrm{test}}$ & F$_{\mathrm{train}}$ \\
\midrule
ResNet-50 & 77.01 & 25.6M & 0.00 & 1.00× & 1.00× \\
\midrule
GMP & 74.29 & 2.56M & 90.00 & 0.10× & 0.51× \\
GraNet & 74.68 & 2.56M & 90.00 & 0.16× & 0.23× \\
Spartan & 75.12 & 2.56M & 90.00 & 0.14× & - \\
AC/DC & 75.83 & 2.56M & 90.00 & 0.18× & 0.58× \\
\textbf{Hyperflux} & 75.28 & 2.54M & 90.11 & 0.15× & 0.51× \\
\midrule
GMP & 70.95 & 1.28M & 95.00 & 0.05× & - \\
GraNet & 72.83 & 1.28M & 95.00 & 0.12× & 0.17× \\
Spartan & 72.92 & 1.28M & 95.00 & 0.08× & - \\
AC/DC & 74.03 & 1.28M & 95.00 & 0.11× & 0.53× \\
\textbf{Hyperflux} & 73.30 & 1.28M & 95.00 & 0.08× & 0.46× \\
\midrule
GMP & 70.62 & 0.90M & 96.50 & - & - \\
GraNet & 71.06 & 0.90M & 96.50 & 0.09× & 0.15× \\
Spartan & 71.13 & 0.90M & 96.50 & - & - \\
AC/DC & 72.50 & 0.90M & 96.50 & - & - \\
\textbf{Hyperflux} & 72.21 & 0.92M & 96.42 & 0.06× & 0.44× \\
\bottomrule
\end{tabular}
\end{wraptable}

To evaluate \textit{Hyperflux} at scale, we apply it to ResNet-50, DeiT-Tiny, and DeiT-Small on the ImageNet-2012 dataset. As shown in Tables~\ref{table:resnet50_imagenet_compute} and \ref{table:comparison_deit}, \textit{Hyperflux} performs competitively against state-of-the-art methods, even at extreme sparsity levels. Notably, for ResNet-50, initializing with pretrained weights improves the accuracy of most baseline methods compared to their reported results, with the sole exception of Spartan, which experiences an accuracy drop of nearly \(1.5\%\).

Focusing first on ResNet-50 (Table~\ref{table:resnet50_imagenet_compute}), \textit{Hyperflux} reaches 72.21\% accuracy at 96.42\% sparsity. This surpasses GMP, GraNet, and Spartan, and remains highly competitive with AC/DC, trailing by only 0.3\%. This performance hierarchy holds consistent at the 90\% and 95\% sparsity targets, where the accuracy gap between \textit{Hyperflux} and AC/DC stays below 0.6\%. To better understand this behavior, we analyzed the weight histograms of the pruned ResNet-50 models, revealing that \textit{Hyperflux} aggressively prunes the convolutional layers; further details on the resulting weight distributions are provided in Appendix~\ref{appendixsecsub:layerwise_sparsity_histo}.

In terms of compute cost (FLOPs), \textit{Hyperflux} proves efficient at inference. For ResNet-50, it reduces test-time compute costs ($F_{\text{test}}$) to 0.15$\times$, 0.08$\times$, and 0.06$\times$ the dense baseline at 90.11\%, 95.00\%, and 96.42\% sparsity, consistently matching or improving upon the efficiency of strong baselines like AC/DC and GraNet. Meanwhile, the training compute costs ($F_{\text{train}}$) of 0.51$\times$, 0.46$\times$, and 0.44$\times$ reflect the theoretical minimum of 1/3 dense FLOPs established in Appendix~\ref{appendixsec:Complexity_analysis}. This is due to the backward pass through the full weight matrix for $t$-value updates, which necessitates a dense matrix multiplication at any sparsity level.

\begin{wraptable}[17]{r}{0.48\textwidth}
\centering
\small
\setlength{\tabcolsep}{3pt}
\renewcommand{\arraystretch}{0.9}
\caption{DeiT-Tiny and DeiT-Small top-1 accuracy and sparsity on ImageNet-2012. We denote by $s$ the sparsity. Baselines from \citet{kuznedelev2023cap}.}
\label{table:comparison_deit}
\begin{tabular}{@{}lcccc@{}}
\toprule
& \multicolumn{2}{c}{\textbf{DeiT-Tiny}} & \multicolumn{2}{c}{\textbf{DeiT-Small}} \\
\cmidrule(lr){2-3} \cmidrule(lr){4-5}
Method & Top-1(\%) & $s$(\%) & Top-1(\%) & $s$(\%) \\
\midrule
DeiT (dense)       & 72.20 & 0.00  & 79.80 & 0.00  \\
\midrule
GM                 & 70.50 & 80.00 & 77.30 & 80.00 \\
CAP                & 71.70 & 80.00 & 78.00 & 80.00 \\
\textbf{Hyperflux} & 71.50 & 80.00 & 77.80 & 80.00 \\
\midrule
SViTE              & 49.70 & 90.00 & 70.10 & 90.00 \\
GM                 & 66.20 & 90.00 & 74.10 & 90.00 \\
CAP                & 67.40 & 90.00 & 75.20 & 90.00 \\
\textbf{Hyperflux} & 67.10 & 90.00 & 74.50 & 90.00 \\
\bottomrule
\end{tabular}
\end{wraptable}

Next we consider the Vision Transformers, reporting results for DeiT-Tiny and DeiT-Small in Table~\ref{table:comparison_deit}. To ensure a strict and fair comparison, we replicate the training recipe of the CAP method \citep{kuznedelev2023cap}, applying the \textit{light1} \citep{steiner2021train} and \textit{deit} \citep{touvron2021training} augmentation schemes with a batch size of 1024. The baseline metrics are sourced directly from the CAP study, which utilizes a 300-epoch gradual pruning schedule. In contrast, \textit{Hyperflux} achieves its results by fine-tuning for only 150 epochs after loading pretrained weights, effectively requiring half the training budget of CAP.

On DeiT-Tiny, \textit{Hyperflux} consistently outperforms GM at all tested sparsity levels and remains within 0.3\% of CAP, despite the significantly reduced training budget. This trend continues on the larger DeiT-Small architecture, where \textit{Hyperflux} surpasses GM at both 80\% and 90\% sparsity while maintaining competitive performance against CAP. Across all Transformer configurations, the accuracy gap compared to CAP never exceeds 0.7\%. This demonstrates that the \textit{Hyperflux} pruning criterion scales effectively to larger, attention-based architectures without necessitating extended training schedules.

\section{Limitations \& Future Work}
\label{section:limitations}

Despite its advantages, Hyperflux possesses several limitations. Our method incurs at least $33\%$ of the computational cost of the dense network (see 
Appendix~\ref{appendixsec:Complexity_analysis}) and introduces additional hyperparameters, such as the scheduler policy and the step size $\eta_t$. While these scheduler parameters generalize across specific task classes (e.g., computer vision), they must still be found. Furthermore, if we choose to use the regrowth stage at the end, the final network sparsity will deviate from its target as regrowth happens after the scheduler is turned off. This can be mitigated (e.g., high learning rates in the beginning of training significantly reduce the amount of regrown weights), but further research is needed to reduce this deviation, or to develop a scheduler that handles it automatically. Additionally, adding large-scale experiments in Section \ref{subsec:experimental_properties} would strengthen our macroscopic explanations and help improve the scheduler. Finally, the manuscript would benefit from evaluating Hyperflux across a broader range of domains, including reinforcement learning and large language models.

\section{Conclusions}
\label{section:discussion}

We presented Hyperflux, a novel $L_0$ regularization method that characterizes weight pruning through the interplay of \textit{flux} and \textit{pressure}, modeling the process as a \textit{continuously evolving system}. This model provides an intuitive mathematical foundation for explaining weight pruning/regrowth decisions and sparsity convergence. Furthermore, we introduce a novel scheduler derived from this framework to accurately achieve target sparsities. Empirically, Hyperflux achieves competitive performance against state-of-the-art pruning methods across CIFAR-10/100 and ImageNet-2012, on both convolutional (ResNet-50, VGG-19) and transformer-based (DeiT) architectures.

\bibliography{hyperflux}
\bibliographystyle{tmlr}

\newpage
\appendix
\section{Analysis}
\label{appendixsec:mathematics}

\subsection{Why Important Weights Generate Stronger Flux}
\label{appendixsecsub:weightimportance_largerflux}
To study the flux of important weights, let us focus on a specific weight \(\omega_i\) in the regime \(t_i \leq 0\), and thus \(\theta_i =0\). For analytical purposes, we define the loss in terms of the effective weights \(\theta_i = \omega_i \cdot H(t_i)\) as \(\mathcal{L}(\theta)\), where \( \mathcal{L}(\theta| \theta_i=0)\) is the loss when \(t_i \leq 0\) and \(\mathcal{L}(\theta| \theta_i=\omega_i)\) is the loss when \(t_i > 0\). We perform a Taylor expansion of \(\mathcal{L}(\theta)\) around \(\theta_i = 0\). By perturbing  \(\theta_i\) by \(\omega_i\) (i.e. by approximating the effect of regrowing the weight), we observe that the first-order term in the expansion is the flux of \(\omega_i\). Formally: 
\begin{equation}
\label{eq:taylor_exp}
\mathcal{L}(\theta| \theta_i=\omega_i)
= \mathcal{L}(\theta| \theta_i=0)
+ \omega_{i}\,\frac{\partial\mathcal{L}(\theta| \theta_i=0)}{\partial \theta_{i}}
+ \frac{1}{2}\,\omega_{i}^{2}\,\frac{\partial^{2}\mathcal{L}(\theta| \theta_i=0)}{\partial \theta_{i}^{2}}
+ O\bigl(\omega_{i}^{3}\bigr).
\end{equation}

Recalling the formula for flux, \(\mathcal{G}_i^-(\omega, \mathcal{T})\), and
neglecting the second and higher-order terms:
\[ 
\mathcal{L}(\theta| \theta_i=0) - 
\mathcal{L}(\theta| \theta_i=\omega_i)
\approx
-\,\omega_{i}\frac{\partial\mathcal{L}(\theta|\theta_i =0)}{\partial \theta_{i}} 
=
\mathcal{G}_i^-(\omega, \mathcal{T}).
\]
Thus we obtain a direct relationship between flux and weight importance: the flux approximates the change in the loss that could be incurred when the weight is regrown. However, this relationship holds only up to neglected higher-order terms, so it should be viewed as a useful approximation rather than an exact law. 

\subsection{Flux Connection To The Hessian}
\label{appendixsecsub:flow_and_hessian}

To relate flux to other importance metrics, specifically the Hessian, we consider the Taylor approximation from (\ref{eq:taylor_exp}) and write:
\[
\mathcal{L}(\theta| \theta_i=0) - 
\mathcal{L}(\theta| \theta_i=\omega_i) = - \left( 
\omega_{i}\,
\frac{\partial\mathcal{L}(\theta| \theta_i=0)}{\partial \theta_{i}} 
+ 
\frac{1}{2}\,\omega_{i}^{2}\,\frac{\partial^{2}\mathcal{L}(\theta| \theta_i=0)}{\partial \theta_{i}^{2}}
\right) 
- 
O\bigl(\omega_{i}^{3}\bigr).
\]
Given that the flux \(\mathcal{G}_i^-(\omega, \mathcal{T}) = -\omega_{i}\,\frac{\partial\mathcal{L}(\theta| \theta_i=0)}{\partial \theta_{i}}\) and neglecting terms of order \(O\bigl(\omega_{i}^{3}\bigr)\) and higher, we obtain:
\[
\mathcal{L}(\theta| \theta_i=0) - 
\mathcal{L}(\theta| \theta_i=\omega_i)
\approx 
\mathcal{G}_i^-(\omega, \mathcal{T}) - \frac{1}{2}\,\omega_{i}^{2}\,
\underbrace{\frac{\partial^{2}\mathcal{L}(\theta| \theta_i=0)}{\partial \theta_{i}^{2}}}_{H_{ii}^{\theta}}
\]
The second term, \(-\frac{1}{2}\,\omega_{i}^{2}\,H_{ii}^{\theta}\), contains the diagonal element, \(H_{ii}^{\theta}\), of the Hessian matrix of the loss function with respect to \(\theta_i\). This shows that our flux metric captures the linear component of the loss change, while the second term captures the quadratic component, which is generally associated with Hessian-based pruning methods like Optimal Brain Damage~\citep{lecun1989optimal}. In Optimal Brain Damage, a weight's saliency is estimated by \(\frac{1}{2} H_{ii} \omega_i^2\), typically under the assumption that the network is at a minimum where first-order gradients are zero. On the other hand, Hyperflux prunes the weights, therefore evaluating the first linear component of the Taylor expansion when the weight is pruned, preventing well converged weights from yielding an approximation close to 0.

\subsection{Relationship to First-Order Saliency Methods}
\label{app:flux_saliency}

The instantaneous flux $\mathcal{G}^-_i(\omega, \mathcal{T})$ is similar to first-order saliency metrics used in prior pruning work. In this section we 
make this connection precise and identify the differences in paradigm, 
granularity, and dynamics.

\paragraph{Mathematical connection.}
The Taylor criterion of~\citet{molchanov2017pruning} approximates the change in loss 
induced by zeroing out a feature map $h_i$ via a first-order Taylor expansion:
\begin{equation}
    \Theta_{TE}(h_i) = \left|\frac{\partial \mathcal{C}}{\partial h_i} \cdot h_i\right|,
\end{equation}
averaged over the spatial positions of the feature map. The connection sensitivity score 
of \citet{lee2019snip} similarly computes $\left|\frac{\partial \mathcal{L}}{\partial c_i} 
\cdot c_i\right|$ with respect to a binary connection mask $c_i$.

The instantaneous flux $\mathcal{G}^-_i(\omega, \mathcal{T}) = \frac{\partial \mathcal{L}}
{\partial \omega_i}\cdot \omega_i$ mathematically has the same form. This similarity is 
not coincidental: all three quantities approximate the change in loss $\Delta\mathcal{L}$ under a first-order 
Taylor expansion. However, as we detail below, 
the mathematical expansion points and the context in which flux is used differ from both \citet{molchanov2017pruning} and ~\citet{lee2019snip}. The differences hold for other first-order methods as well. 

\paragraph{(i) Application to pruned vs. active weights.}
Standard first-order methods, such as the Taylor criterion~\citep{molchanov2017pruning} and SNIP~\citep{lee2019snip}, compute importance scores exclusively on active weights to determine which to remove. Mathematically, they perform a Taylor expansion around the current active parameter (expansion point $a = \omega_i$) and evaluate the estimated loss when the weight is zero. This dynamic introduces a major issue at convergence: if an active weight is important but close to a local minimum, its gradient approaches zero (\(\frac{\partial \mathcal{L}}{\partial \omega_i}\big|_{\omega_i} \approx 0\)). Consequently, the first-order approximation falsely estimates a near-zero change in loss, risking the removal of critical features. 

Hyperflux bypasses this blind spot by altering the Taylor expansion paradigm. Instead of expanding around the active weight, Hyperflux computes flux specifically on \textit{pruned} weights (i.e., when $t_i \leq 0$ and the effective weight $\theta_i = 0$). It sets the expansion point at the pruned state ($a = 0$) and evaluates the estimated change in loss if the weight were to be regrown to its dense value ($x = \omega_i$). By evaluating the gradient at $0$, far from the near optimal, zero gradient region, the local loss landscape is steep. If the weight is important, the gradient evaluated at $0$ (\(\frac{\partial \mathcal{L}}{\partial \omega_i}\big|_0\)) will be far from zero, accurately capturing the mathematical demand for the weight and driving its regrowth. This formalizes the insight that one does not fully know the value of a parameter until it is removed.

\paragraph{(ii) Continuous dynamics vs. greedy stop-and-sort.}
~\citet{molchanov2017pruning} implement pruning as an explicit iterative loop: training is paused, active units are ranked, the lowest ranked are removed, and fine-tuning resumes. This \textit{stop-and-sort} procedure necessitates rigid, manual scheduling of discrete pruning stages. Similarly, SNIP~\citep{lee2019snip} is bounded by a single, rigid ranking at initialization. In contrast, Hyperflux uses an \(L_0\)-like regularization to continuously prune candidate weights and evaluate if they should be regrown via flux. This allows the model to smoothly navigate towards a joint optimum of weights and sparsity, replacing manual schedules with mathematical optimization.

\paragraph{(iii) Aggregated vs. instantaneous evaluation.}
Both the Taylor criterion and SNIP evaluate weight importance at a single topology. A score computed at topology $\mathcal{T}^k$ 
reflects only the network's state at that moment. In Hyperflux, the aggregated flux $\mathcal{G}^-_i(\omega, \mathcal{T}^{1 \to K})$ accumulates the gradient signal over a \textit{family of topologies} visited during training. This means a weight's importance is assessed across many different network configurations, providing a more robust metric. A weight deemed unimportant at topology $\mathcal{T}^k$, and therefore pruned, can be regrown at a later topology $\mathcal{T}^{k'}$ if accumulated evidence from the 
loss landscape demands it, something that is impossible in both \citet{lee2019snip} and 
\citet{molchanov2017pruning}.

\subsection{Comparison with Differentiable Mask Methods}
\label{app:diff_mask}

Differentiable mask methods share with Hyperflux the goal of jointly optimizing weights 
and a pruning mask through gradient-based optimization, with per-weight (or per-group) 
learnable parameters governing sparsity. A key distinction of Hyperflux is that it 
introduces an explicit pressure scheduler \textsc{Sched}$(d_e, e)$ that adapts pruning 
pressure based on current network density and training epoch, enabling principled 
sparsity targeting. Furthermore, Hyperflux maintains a non-zero gradient signal through 
pruned weights via a constant Straight-Through Estimator ($\text{STE}_H(t_i) = 1$) 
throughout the entire training process, meaning any pruned weight can be regrown at any 
point if the loss demands it. None of the methods below share both of these properties.

Continuous Sparsification~\citep{savarese2020winningcontinuous} freezes pruning decisions 
in later training stages as its global temperature $\beta(t) \to \infty$ causes gradients 
to vanish, a limitation the authors themselves acknowledge, making weight regrowth 
impossible once the mask has hardened. Furthermore, the temperature schedule is global, 
hardening all weights at the same rate regardless of their individual importance, whereas 
in Hyperflux pruning and regrowth decisions are driven locally by per-weight flux and 
pressure at every training step.

OptG~\citep{zhang2022optg} shares with Hyperflux the motivation of accumulating gradient 
signals for both active and pruned weights. However, it alternates between frozen-weight 
mask optimization and frozen-mask weight optimization, meaning weights and topology are 
never co-optimized simultaneously as they are in Hyperflux. The final topology is 
determined by a global top-$k$ sort over accumulated scores, whereas in Hyperflux 
topology emerges continuously from the local interaction between per-weight flux and 
global pressure at every training step, with no explicit ranking required.

Stochastic $L_0$ Regularization~\citep{louizos2018l0} shares with Hyperflux the use of 
per-weight learnable parameters and an $L_0$ penalty on active connections. However, in $L_0$ regularization, sparsity emerges as a side effect of minimizing the expected gate activation, and cannot be controlled precisely, leading to expensive grid searches over the initial regularization. Furthermore, a large regularization required for high sparsity may prune the network rapidly, leading to accuracy degradation or collapse. Hyperflux instead controls regularization via a scheduler \textsc{Sched}$(d_e, e)$, allowing our method to start with a small regularization initially, which preserves accuracy, and gradually increase it to steer the network towards the desired sparsity, solving both issues mentioned above. 

\subsection{Pruning process model caveats}
\label{app:PruningProcessDisc}

While the flux and pressure model presented in Section~\ref{section:Hyperflux} provides an intuitive and effective framework for understanding pruning, it relies on several simplifying assumptions. To provide a rigorous evaluation of our method, we highlight the potential gaps in this model and discuss why our criteria serve as heuristic approximations rather than absolute oracles of weight utility.

\textbf{i) Why $\widehat{\mathcal{G}}_i^+$ prunes plausible evaluation candidates.} Consider the criteria for pruning candidates governed by $\widehat{\mathcal{G}}_i^+$. Our model assumes that a weight moving away from zero is a bad candidate for evaluation since it's most likely important, while a weight drifting toward zero is a good candidate for evaluation since it's probably unimportant. However, the true utility of a parameter is not strictly monotonic with respect to its gradient trajectory. For instance, a weight with $\widehat{\mathcal{G}}_i^+ < 0$ may indeed be unimportant, but it could equally be converging to a smaller-but-useful local optimum. Similarly, a weight with $\widehat{\mathcal{G}}_i^+ \approx 0$ might not be useless; it could be a highly significant, fully converged parameter residing in a flat region of the loss landscape. Furthermore, a weight satisfying $0 < \widehat{\mathcal{G}}_i^+ < \frac{\gamma}{d}$ may be drifting toward an optimal value, yet its gradient is simply too weak to overcome the global pressure. In this light, pruning does not definitively measure insignificance; rather, it acts as a ``good enough'' probabilistic filter that forces weights with ambiguous trajectories into a pruned state where their true importance can be tested via flux.

\textbf{ii) Gradient tendency versus observed magnitude change.} Our mathematical formulation defines $\mathcal{G}_i^+$ as the \textit{tendency} of the weight's magnitude to change, rather than the observed change itself. This distinction is necessary due to the complex dynamics of network optimization. For example, standard $L_2$ regularization (weight decay) acts directly on the weights and may perfectly cancel out a positive loss gradient. In such a scenario, the weight's actual magnitude remains stagnant. However, because the presence parameter $t_i$ follows the task-specific gradient tendency rather than the net weight update, $t_i$ will correctly continue to increase. Moreover, when a weight is masked ($t_i \leq 0$), its magnitude is strictly zero and cannot change, yet this underlying tendency is still actively measured and accumulated as flux. 

\textbf{iii) Finite iterations for pruning/regrowth decisions and weight decay.} While the interplay of flux and pressure provides a mechanism for continuous evaluation, in practice, this topological exploration operates within a finite window due to lack of infinite time. When a weight remains pruned (\(t_i \leq 0\)), its gradient from the task loss is zero. However, standard \(L_2\) regularization (weight decay) continues to penalize the underlying magnitude \(\omega_i\), decaying it toward zero. As \(\omega_i \to 0\), the resulting flux \(\mathcal{G}_i^- \to 0\), eventually rendering the weight mathematically incapable of overcoming the pressure \(\frac{\gamma}{d}\). Furthermore, weights that remain pruned for extended periods accumulate large negative \(t_i\) values, creating an inertia barrier against regrowth. We investigated explicit countermeasures to enforce infinite evaluation, such as applying strict lower bounds to \(t_i\) and masking weight decay for pruned weights, but observed no empirical improvement in final network accuracy or sparsity scaling. Consequently, this dynamic acts as a natural annealing process: early in training, the network freely explores topologies, whereas in later stages, deeply pruned weights settle into a permanently dormant state, stabilizing the active subnetwork for final convergence.

\textbf{iv) Limitations of local linear approximations.} The theoretical link between flux and weight importance relies on a first-order Taylor expansion of the loss function, as detailed in Appendix~\ref{appendixsecsub:weightimportance_largerflux}. Because the Taylor expansion is only a local linear approximation, it may fail to accurately capture the highly non-linear change in the loss when a weight completely disappears or is regrown. While this means flux might occasionally be a poor approximation for the true loss landscape, this limitation is not unique to Hyperflux; it is a fundamental constraint shared by all classical first-order pruning criteria.






\section{Ablation Studies}
\label{appendixsec:ablation_studies}

\subsection{Factors influencing flux \(\mathcal{G}_i^-(\omega, \mathcal{T})\)}
\label{appendixsecsub:factors-influencing-flux}

\begin{wrapfigure}{r}{0.48\textwidth}
    \centering
    \includegraphics[width=\linewidth]{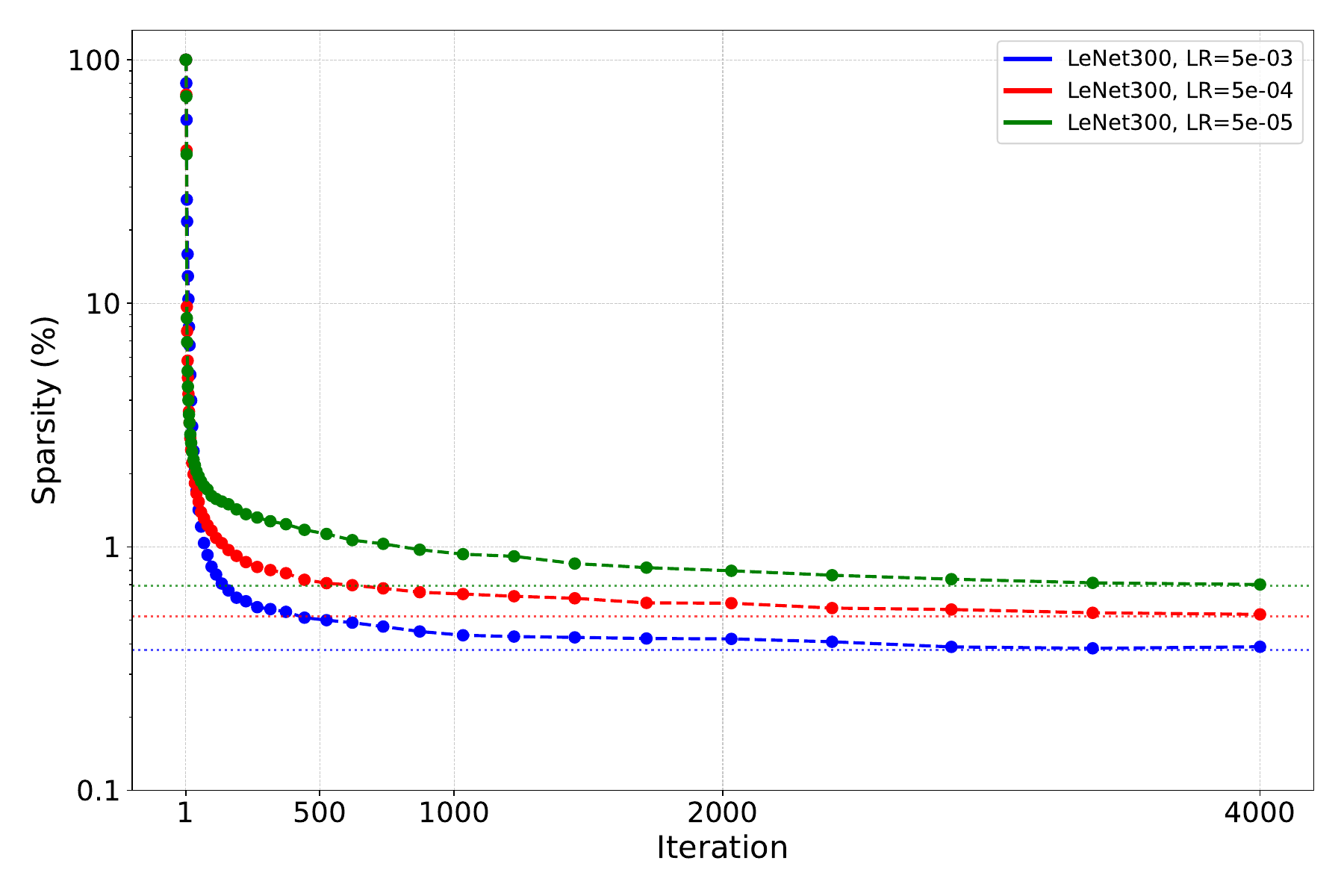}
    \caption{MNIST convergence for constant \(\theta = 1\) for different learning rates}
    \label{fig:mnist_convergence_ablation}
\end{wrapfigure}

We begin by analyzing how the flux value \(\mathcal{G}_i^-(\omega, \mathcal{T})\) is influenced by factors other than \(\eta_t\), the learning rate on presence parameters. Our findings from Section~\ref{subsec:experimental_properties} suggest that weight learning affects the behavior of flux by changing the final convergence point a network will reach for the same constant pressure \(\gamma\). We study this effect in the case of LeNet-300. We run the network for 1000 epochs for three different learning rates of \(0.005, 0.0005\) and \(0.00005\), with no schedulers used and the same constant \(\gamma\). 
Our findings are summarized in Figure~\ref{fig:mnist_convergence_ablation}, which shows that increasing \(\eta_{\omega}\), the weights learning rate, leads to smaller fluxes and convergence at higher sparsities.  

Given the impact of \(\eta_{\omega}\) on network convergence, we study the influence of high and low learning rates on our pruning and regrowth phases. In our experiments, we study three setups on ResNet-50 with CIFAR-10. In the first two experiments, we study how constant learning rates across the entire pruning and regrowth process affect sparsity and regrowth. We choose a high learning rate of \(0.01\) and a low learning rate of \(0.0001\). 

\begin{figure}[!htbp]
  \centering
  \begin{subfigure}[b]{0.48\textwidth}
    \centering
    \includegraphics[width=\linewidth]{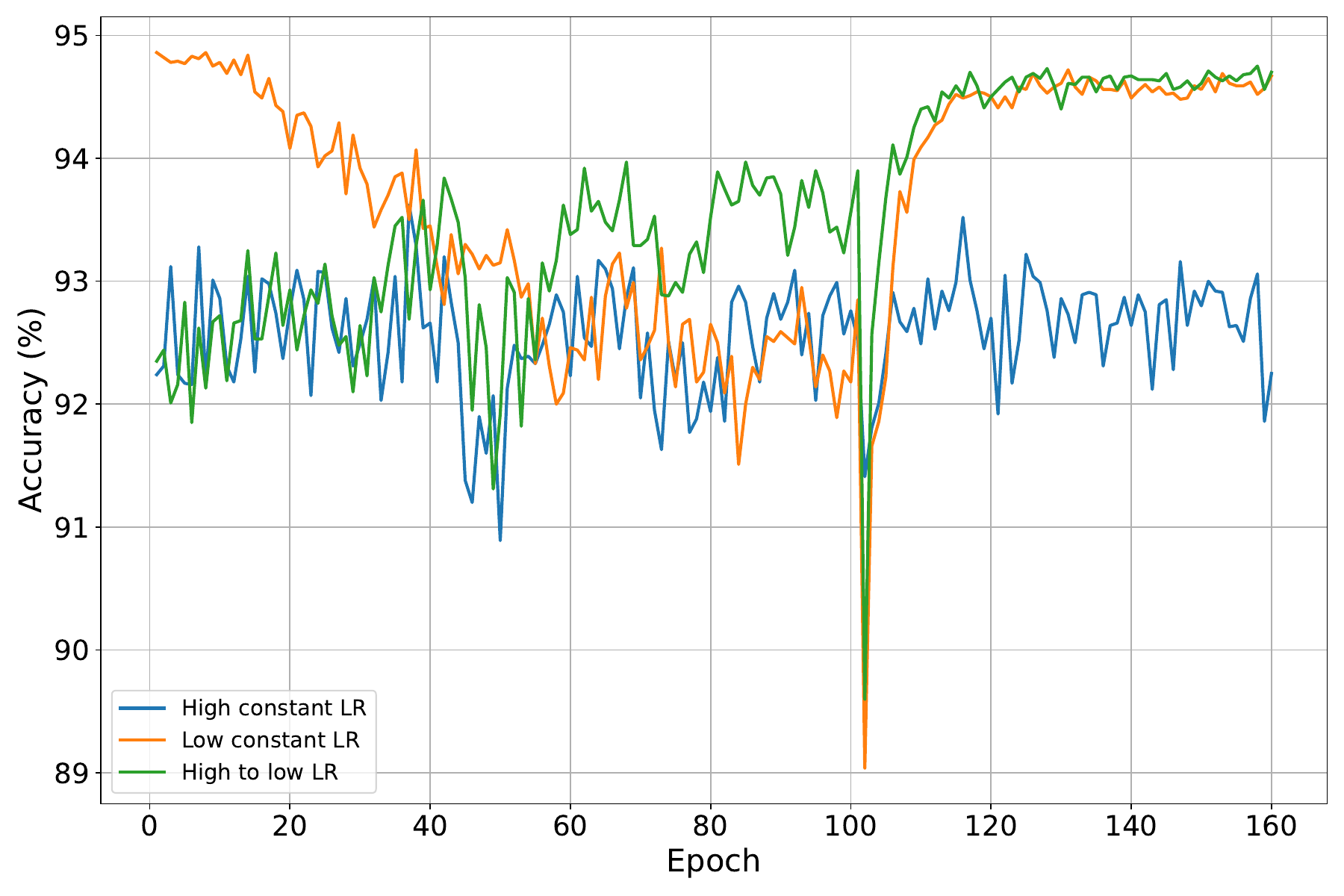}
    \caption{Pruning accuracy vs.\ LR.}
    \label{fig:flux_lrs_accuracy}
  \end{subfigure}
  \hfill
  \begin{subfigure}[b]{0.48\textwidth}
    \centering
    \includegraphics[width=\linewidth]{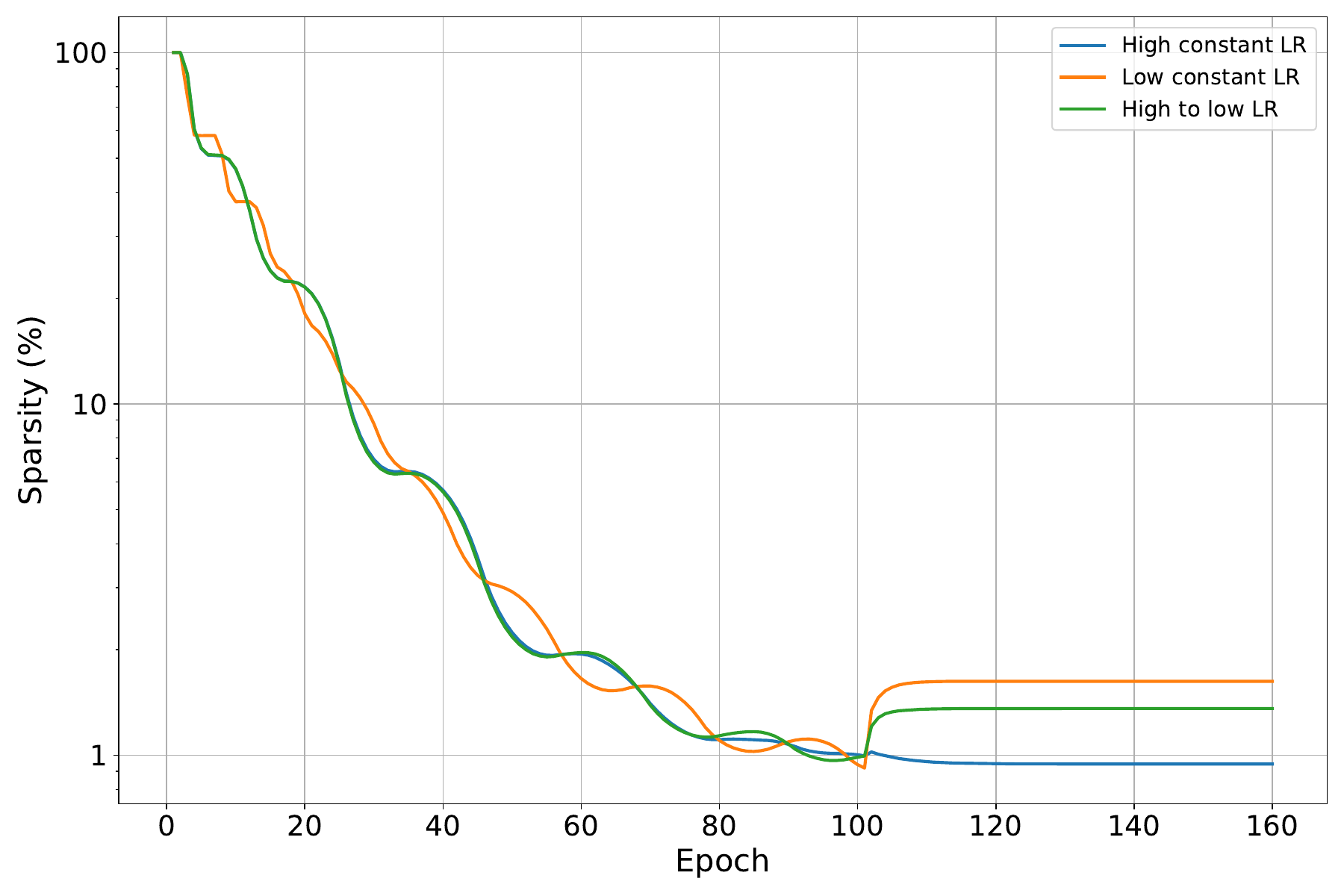}
    \caption{Sparsity vs.\ LR.}
    \label{fig:flux_lrs_sparsity}
  \end{subfigure}

  \caption{The impact of the weights learning rate on pruning accuracy (left) and achieved sparsity (right).}
  \label{fig:appendix_weights_lr_for_flux}
\end{figure}

For our third experiment, we start with the high learning rate which is then decayed using cosine annealing to a low learning rate until the end of regrowth. For all three studies we let our scheduler guide the network towards the same sparsity rate of \(1\%\). However, we observe significant differences in the regrowth stage. For the first experiment, regrowth does not occur at all, with more weights being pruned even after the pressure is set to 0, while for the low learning rate, the performance initially degrades, but is followed by a substantial regrowth stage where the number of remaining parameters increases by \(60\%\). For the third experiment, performance does not degrade as much as for the low learning rate and the regrowth is done in a more controlled way, experiencing an increase in remaining parameters of \(35\%\). The results are illustrated in Figure~\ref{fig:appendix_weights_lr_for_flux}. 
Lastly, we study how weight flux is affected by weight decay. Being directly applied on the weights, weight decay acts on both pruned and present weights. If a weight has been pruned in the first epochs on the training, weight decay will keep making it smaller and smaller, in this way diminishing its flux. We run similar experiments to the ones before, with a learning rate of \(0.01\), decayed during training to \(0.0001\), both with and without the standard weight decay. As expected, we observe in Figure~\ref{fig:appendix_with_decay_and_without} that regrowth without weight decay is more ample. We run this experiment five times, and note that each time the pattern illustrated in the figure remains consistent.

\subsection{Weights \(\eta_{\omega}\) and pruning}
\label{appendixsecsub:weightslr-and-pruning}

Given the large impact \(\eta_{\omega}\) has on flux, we explore its implications for producing an optimal pruning setup for Hyperflux. We run three experimental setups on ResNet-50 CIFAR-10 similar to the ones before. For each one of them, we select a starting learning rate, which is then decayed during training to \(0.0001\) to ensure convergence. For this setup, we run experiments using \(\eta_\omega = {0.1, 0.01, 0.0001}\). We analyze the results from the perspective of accuracy after pruning, noise, regrowth, and final accuracy. We find that the third setup is the most effective for Hyperflux. 

\begin{figure}[!htbp]
  \centering
  \begin{subfigure}[b]{0.48\textwidth}
    \centering
    \includegraphics[width=\linewidth]{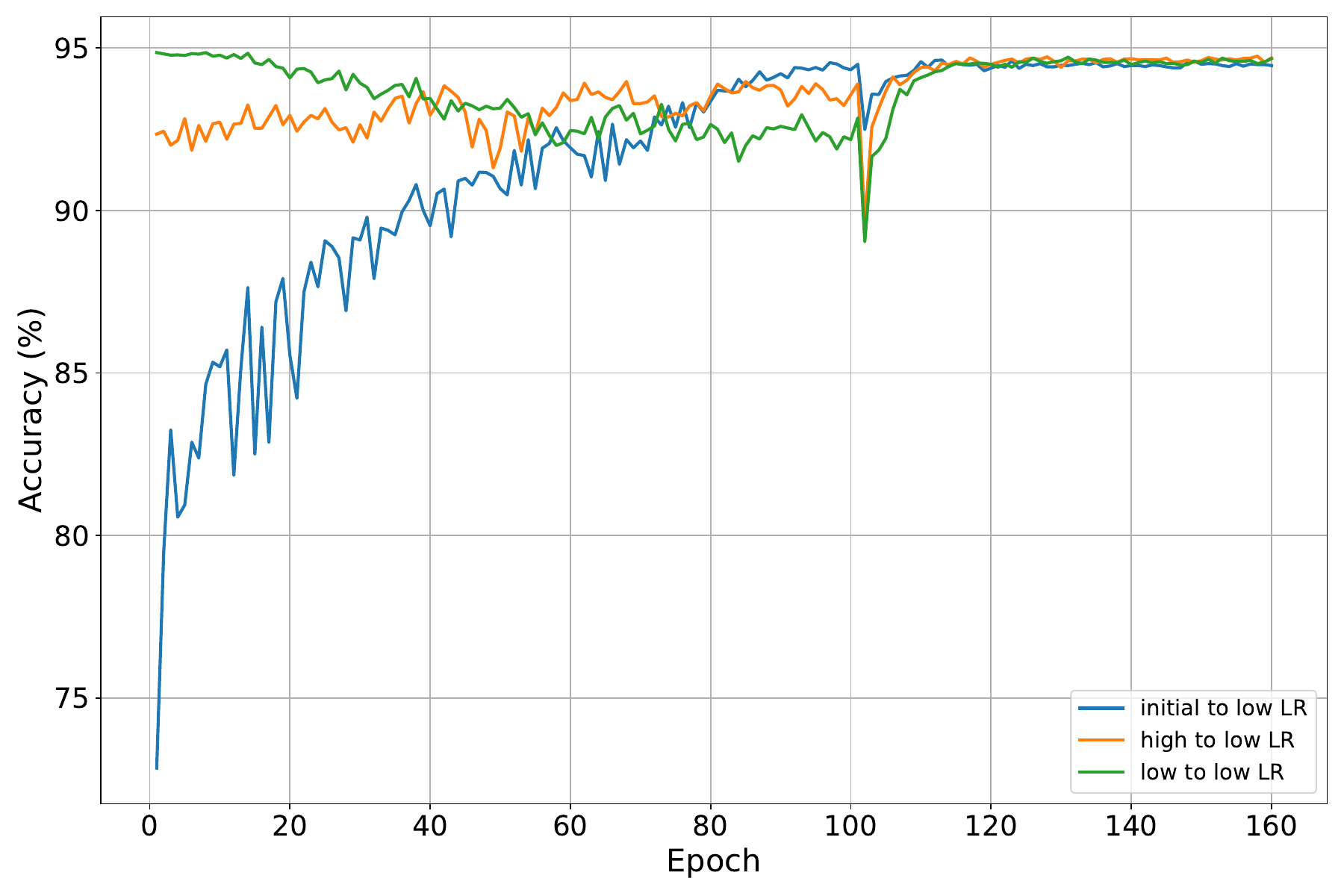}
    \caption{Training accuracy over epochs.}
    \label{fig:sub1_2}
  \end{subfigure}
  \hfill
  \begin{subfigure}[b]{0.48\textwidth}
    \centering
    \includegraphics[width=\linewidth]{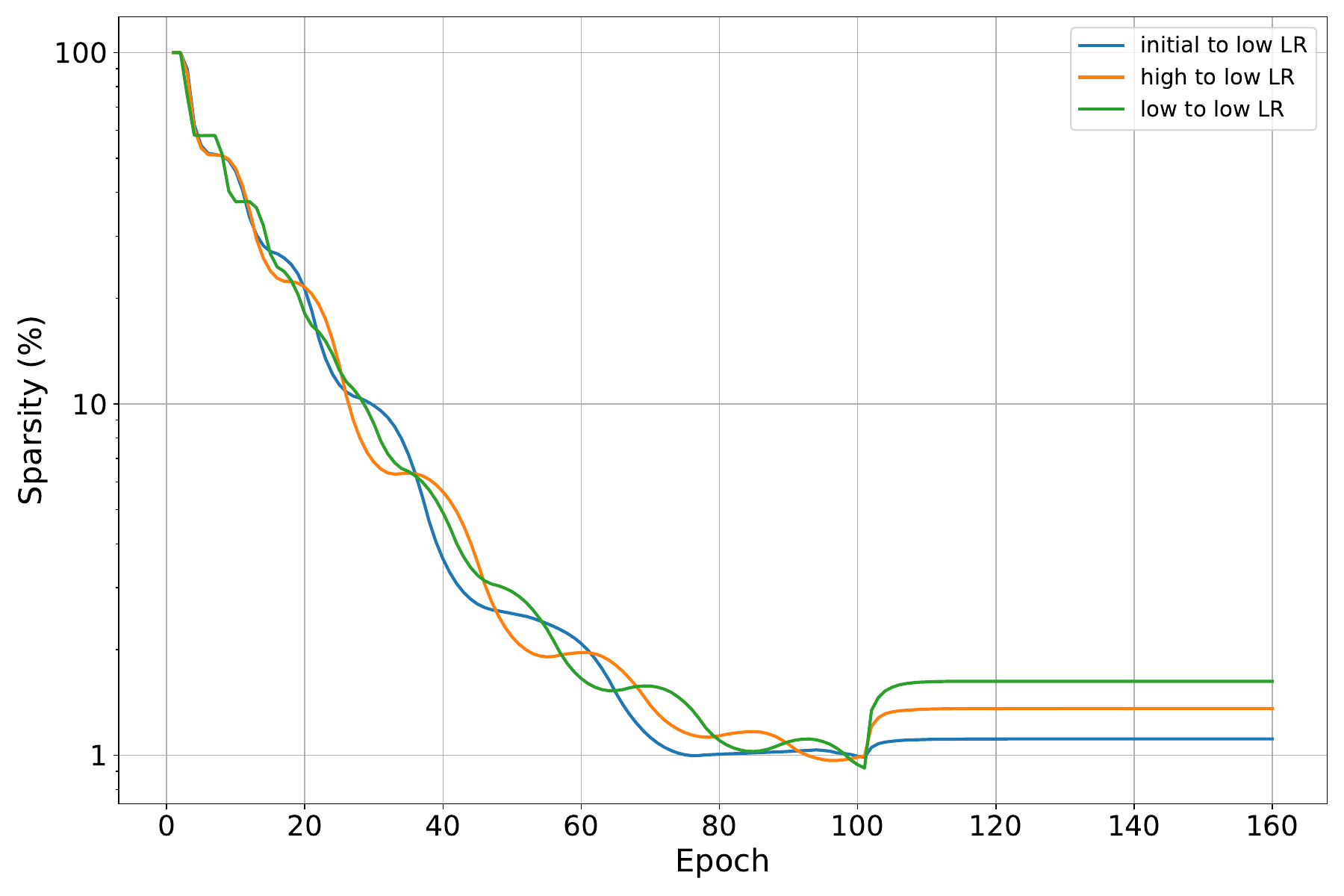}
    \caption{Training sparsity over epochs.}
    \label{fig:sub2_2}
  \end{subfigure}

  \caption{Training evolution for different learning rate configurations.}
  \label{fig:appendix_lr_for_training}
\end{figure}

Each of the four aspects studied is affected by the learning rate. Noise during pruning grows with the initial learning rate. Accuracy at the end of pruning is lowest for small learning rates and highest for large ones. Final accuracy is also highest for larger learning rates. Conversely, the extent of regrowth decreases as the learning rate grows. These trends are visible in Figure~\ref{fig:appendix_lr_for_training}.

\subsection{\(\eta_t\) values and regrowth }
\label{appendixsecsub:tlr-and-regrowth}
\begin{figure}[!htbp]
  \centering
  \begin{subfigure}[b]{0.48\textwidth}
    \centering
    \includegraphics[width=\linewidth]{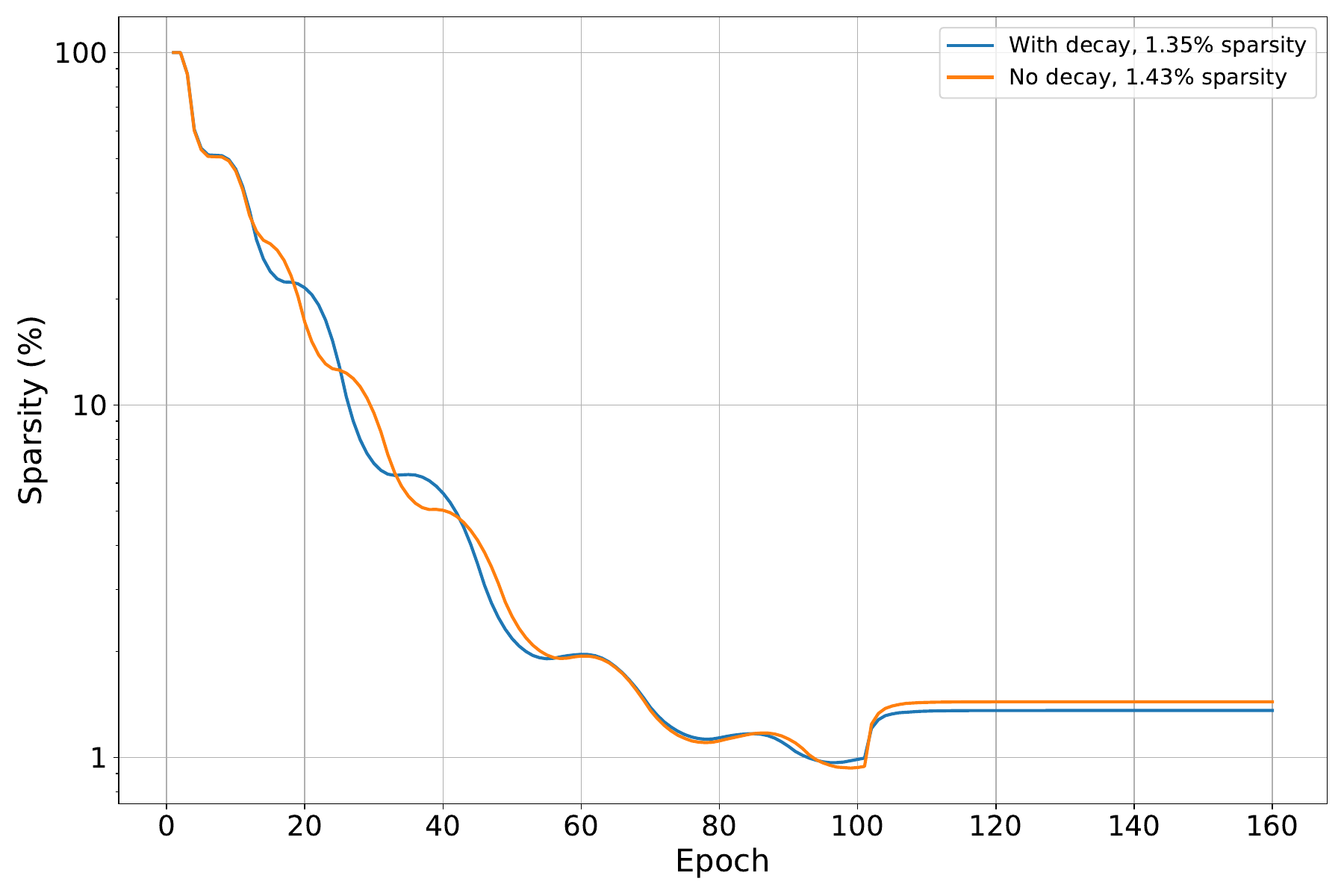}
    \caption{Effect of weight decay on the regrowth process}
    \label{fig:appendix_with_decay_and_without}
  \end{subfigure}
  \hfill
  \begin{subfigure}[b]{0.48\textwidth}
      \centering
    \includegraphics[width=\textwidth]{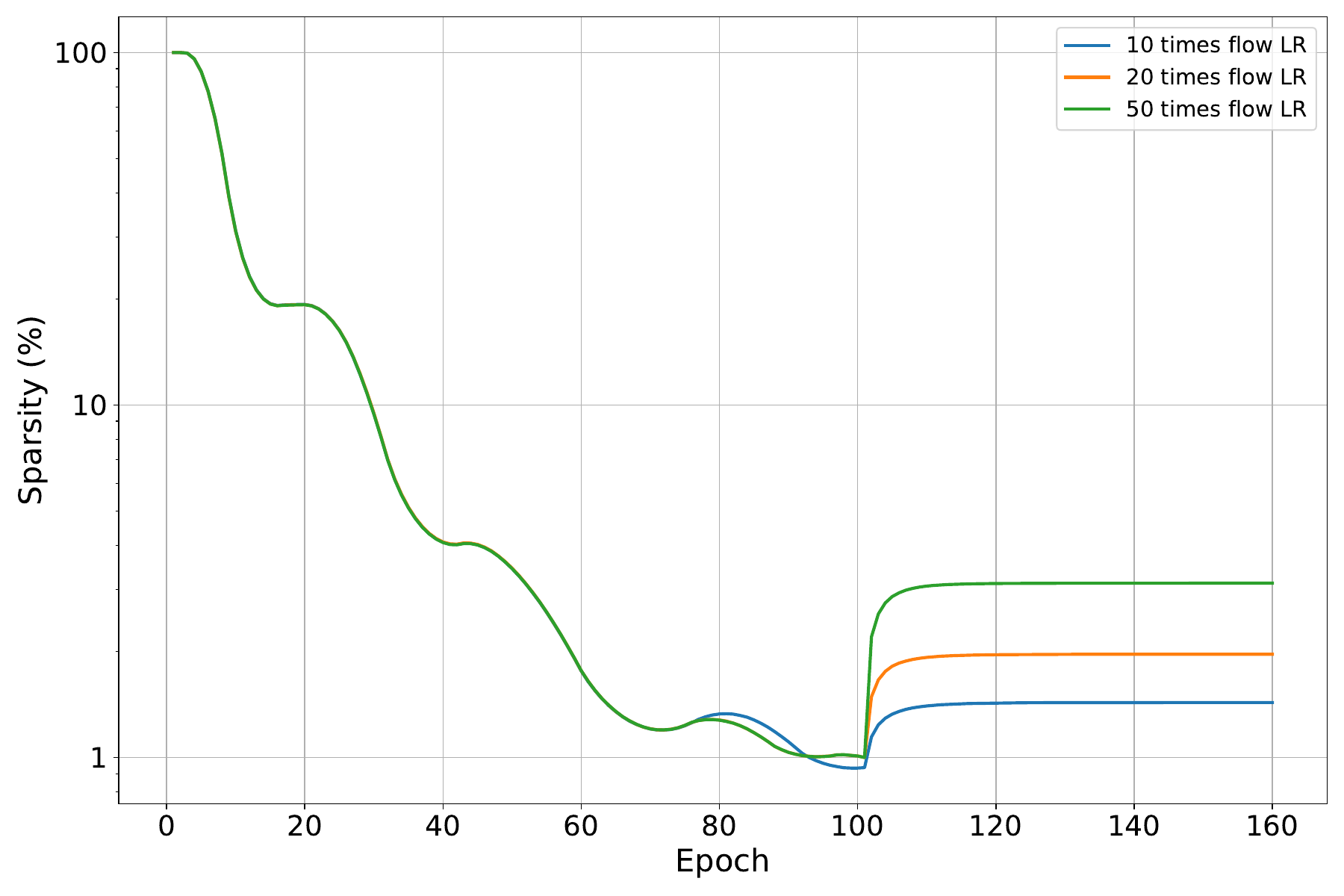}
    \caption{How differently scaled \(\eta_t\) affect regrowth}
    \label{fig:appendix_theta_scaler_vs_regrowth}
  \end{subfigure}

  \caption{Factors affecting regrowth.}
  \label{fig:appendix_lr_for_training_1}
\end{figure}

We analyze regrowth behavior for several values of \(\eta_t\). At regrowth stage, we scale \(\eta_t\) by \(5, 10, 20, 30\) for VGG-19 on CIFAR-100 to observe the behavior of regrowth stage. Our findings are summarized in Figure~\ref{fig:appendix_theta_scaler_vs_regrowth}. As \(\eta_t\) increases so does the number of regrown weights. However, we note that after a point, generally about an increase of \(50\%\) in remaining parameters, the effects of regrowth start to diminish and introduce noise in the performance, while also regrowing more weights.

\section{Extended experiments}
\label{appendixsec:extended_experiments}

\subsection{Layerwise sparsity levels \& Weight Histograms}
\label{appendixsecsub:layerwise_sparsity_histo}
\begin{figure*}[htbp]
    \centering
    \includegraphics[width=1.0\columnwidth]{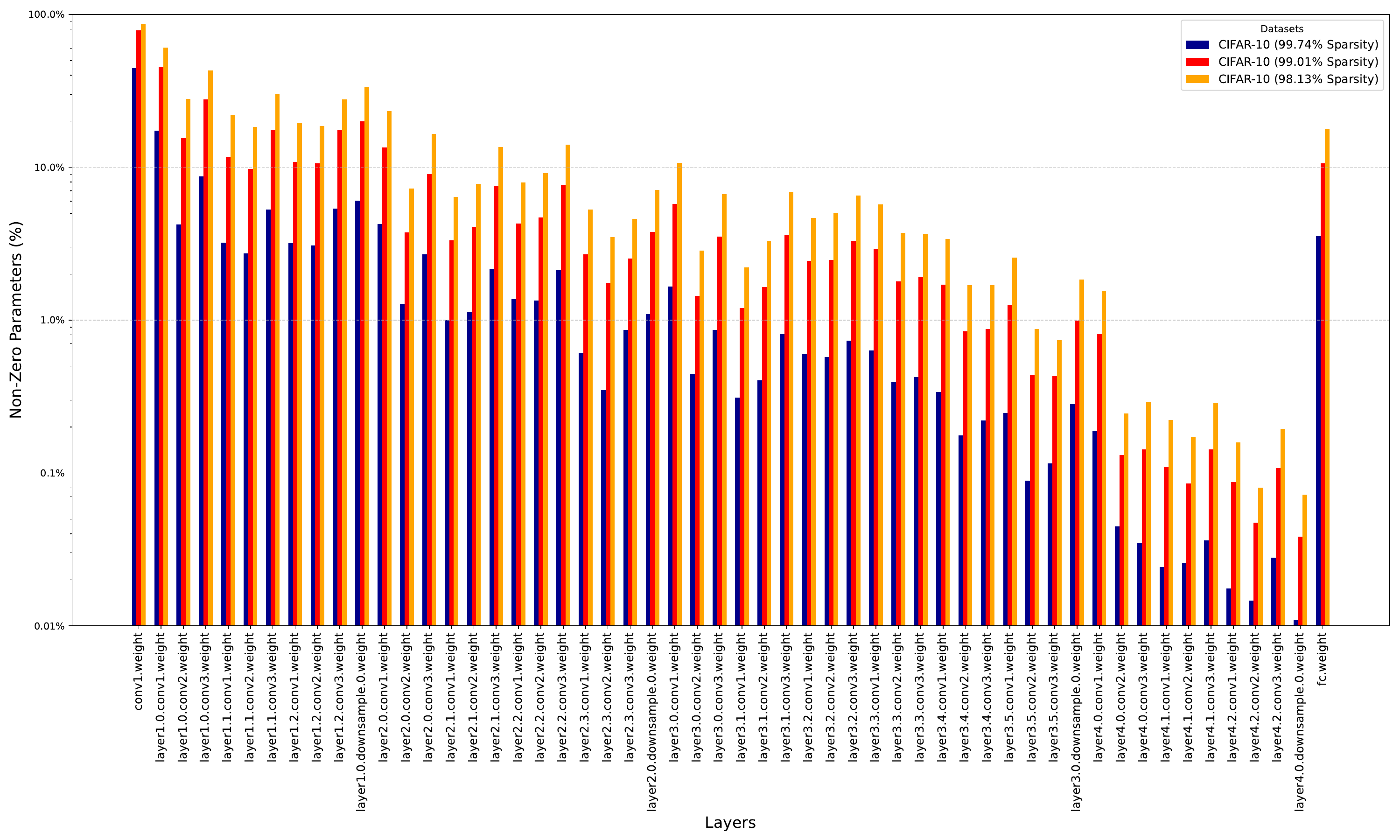}
    \caption{Per-layer sparsity for ResNet-50 CIFAR-10. We present 3 levels of sparsity: \(99.74\%,99.01\%\) and \(98.13\%\).}
    \label{fig:per_sparsity_layer}
\end{figure*}

\begin{figure}[!htbp]
  \centering

  \begin{subfigure}[b]{\textwidth}
    \centering
    \includegraphics[width=\linewidth]{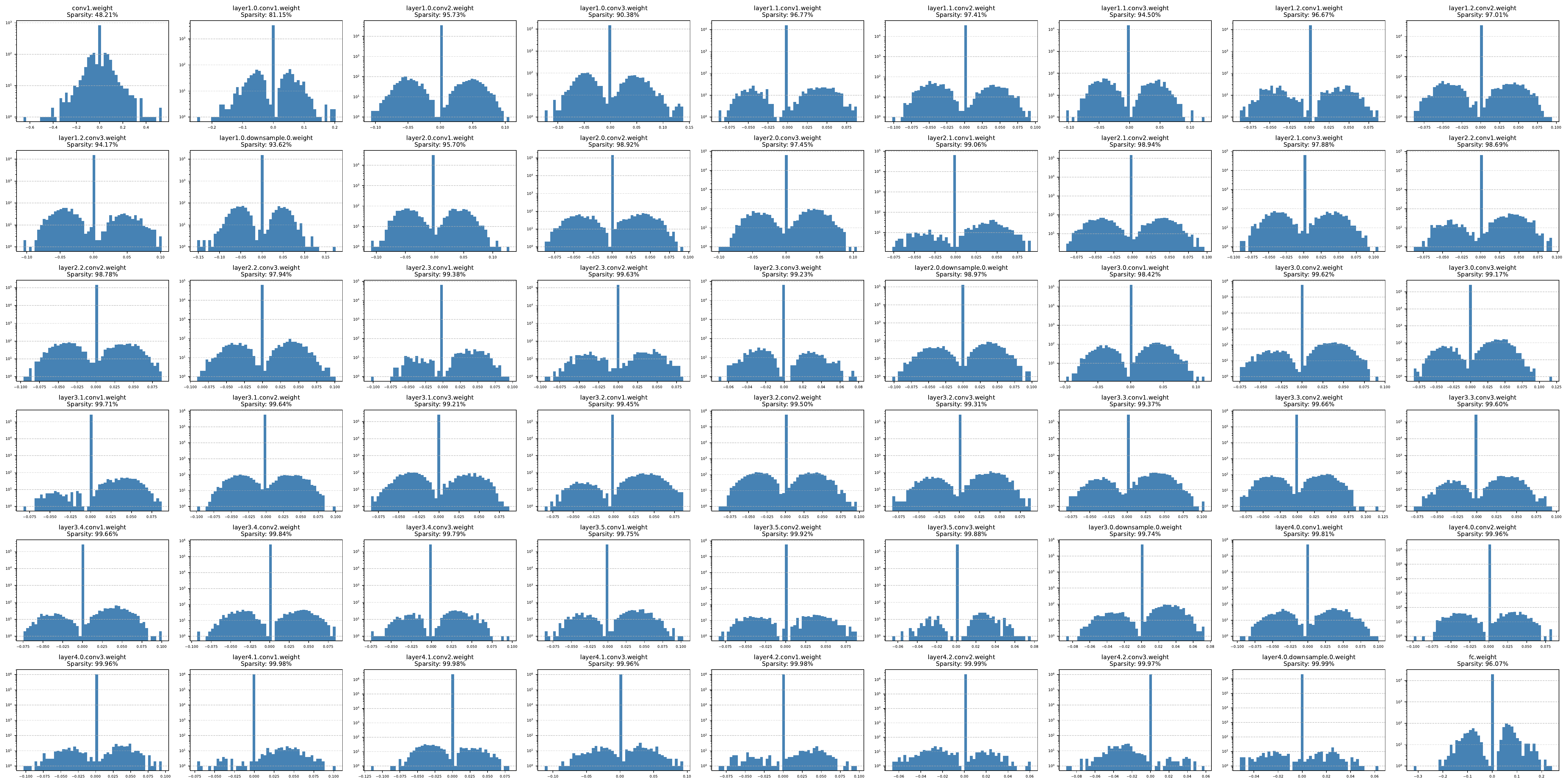}
    \caption{99.74\% sparsity (92.81\% acc).}
    \label{fig:resnet50_CIFAR10_sparse_low}
  \end{subfigure}

  \vspace{1em}

  \begin{subfigure}[b]{\textwidth}
    \centering
    \includegraphics[width=\linewidth]{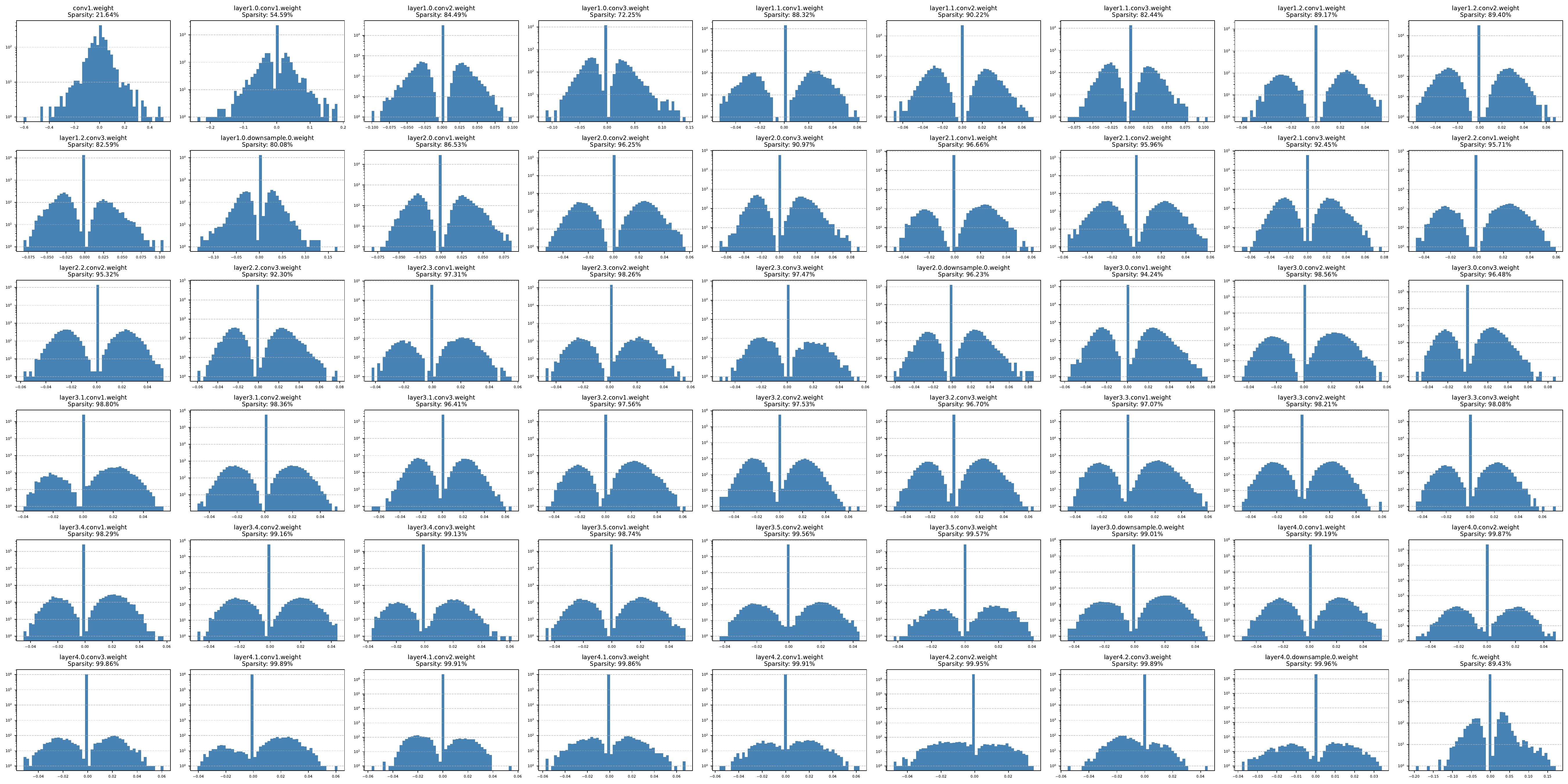}
    \caption{99.1\% sparsity (94.44\% acc).}
    \label{fig:resnet50_CIFAR10_sparse_high}
  \end{subfigure}

  \caption{Weight‐value histograms of ResNet‑50 on CIFAR‑10 at two different sparsity levels. Note how the weight distribution reshapes as sparsity increases.}
  \label{fig:resnet50_CIFAR10_combined_histograms}
\end{figure}

In this section, we examine the layer-wise sparsity observed for ResNet-50 on CIFAR-10 across the following pruning rates: 99.74\%, 99.01\%, and 98.13\%. As illustrated in Figure~\ref{fig:per_sparsity_layer}, the overall sparsity hierarchy is maintained, displaying a decreasing trend in sparsity from the initial layers down to the final layer, where this pattern is interrupted. We hypothesize that earlier layers retain more weights due to their critical role in feature extraction, while deeper layers can sustain higher levels of pruning without significantly impacting overall performance. Notably, the penultimate layer experiences the highest degree of pruning, which means that it contains higher redundancy or less critical weights for performance. Furthermore, by analyzing the weight histograms for ResNet-50 with sparsity levels of 99.01\% and 99.74\% in Figure~\ref{fig:resnet50_CIFAR10_combined_histograms}, we observe the influence of sparsity on the weight distributions. High sparsity levels significantly alter weight distributions, demonstrating that extreme pruning not only reduces the number of active weights but also changes the underlying weight dynamics within the network. 

\begin{figure}[!htbp]
  \centering

  \begin{subfigure}[b]{\textwidth}
    \centering
    \includegraphics[width=\linewidth]{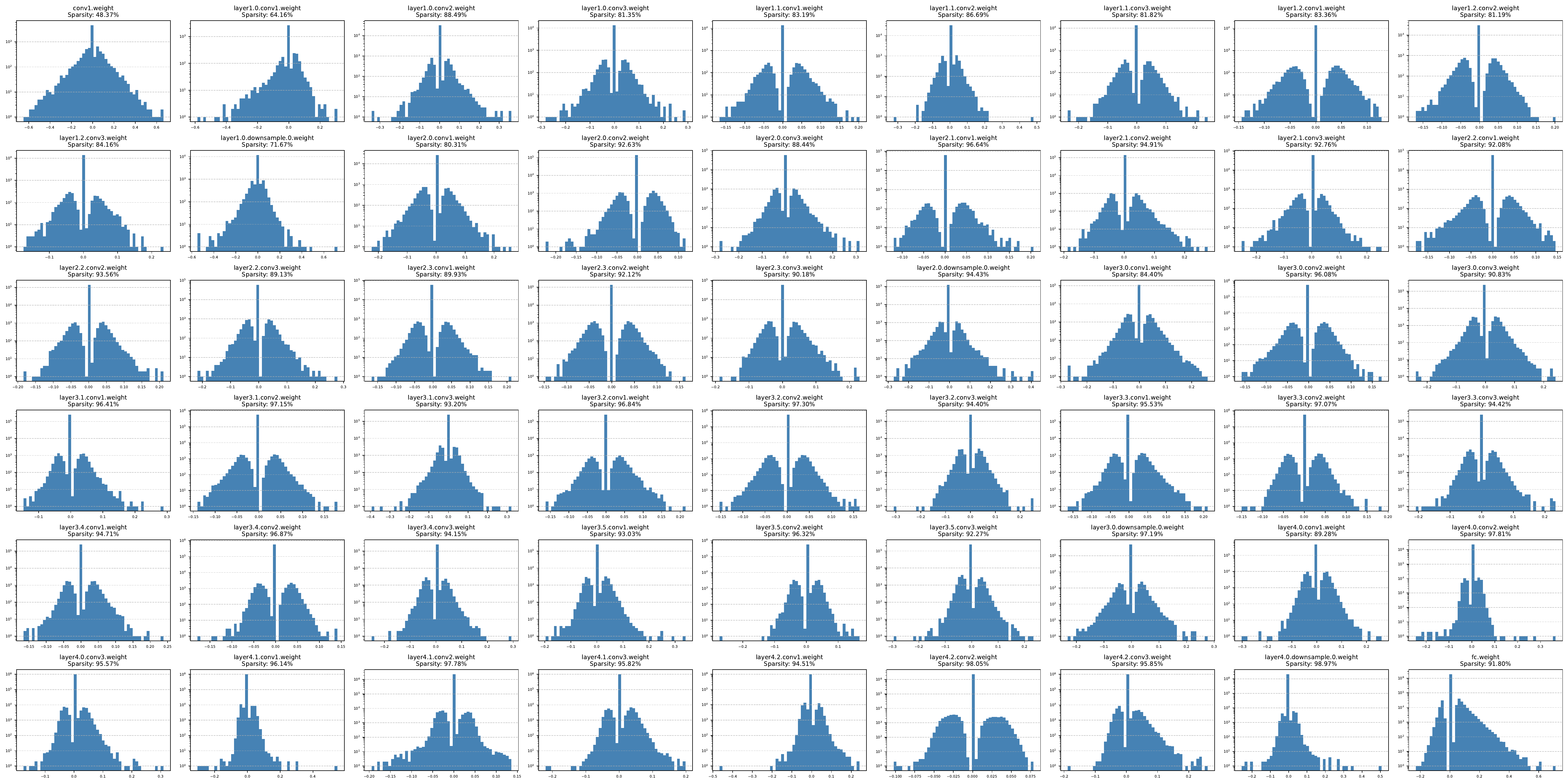}
    \caption{Pruning phase (66.13\% acc).}
    \label{fig:resnet50_prune}
  \end{subfigure}

  \vspace{1em}

  \begin{subfigure}[b]{\textwidth}
    \centering
    \includegraphics[width=\linewidth]{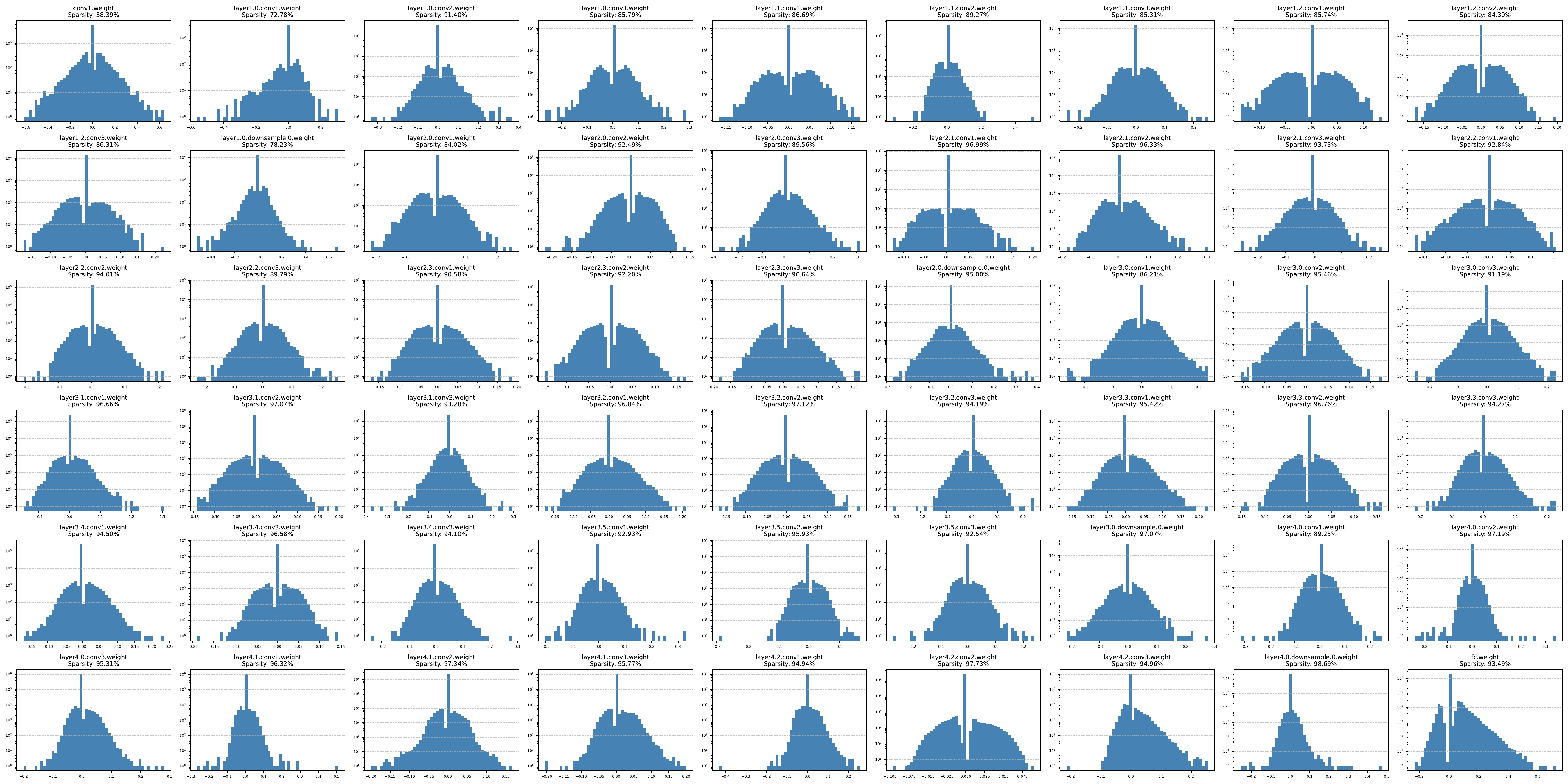}
    \caption{Regrowth phase (72.40\% acc).}
    \label{fig:resnet50_regrow}
  \end{subfigure}

  \caption{Weight histograms of ResNet‑50 on ImageNet during two different phases at 95.77\% sparsity.}
  \label{fig:resnet50_imagenet_combined_histograms}
\end{figure}

The histograms in Figure~\ref{fig:resnet50_imagenet_combined_histograms} illustrate the differences in weight distributions between the pruning and regrowth stages on ImageNet with ResNet-50 at approximately 4.23\% remaining weights. In the pruning stage, weights are more evenly distributed across the range of \([-0.4, 0.4]\), with a noticeable dip near zero, reflecting the removal of low-magnitude weights. In contrast, during the regrowth stage the weight distribution shifts significantly, showing a sharp clustering of weights around zero, indicating the reactivation of low-magnitude weights during this process. This change in distribution correlates with a notable performance gap: the regrowth stage achieves 72.4\% accuracy, while the pruning stage reaches only 66.13\%. We attribute this to the regrowth dynamic: pruning removes low-magnitude weights, and the regrowth phase then reactivates only those that most improve performance.

\subsection{Implicit regrowth and $t$ gradients}
\label{appendixsecsub:implicit-regrowth}
\begin{figure}[htbp]
    \centering
    \includegraphics[width=1.0\columnwidth]{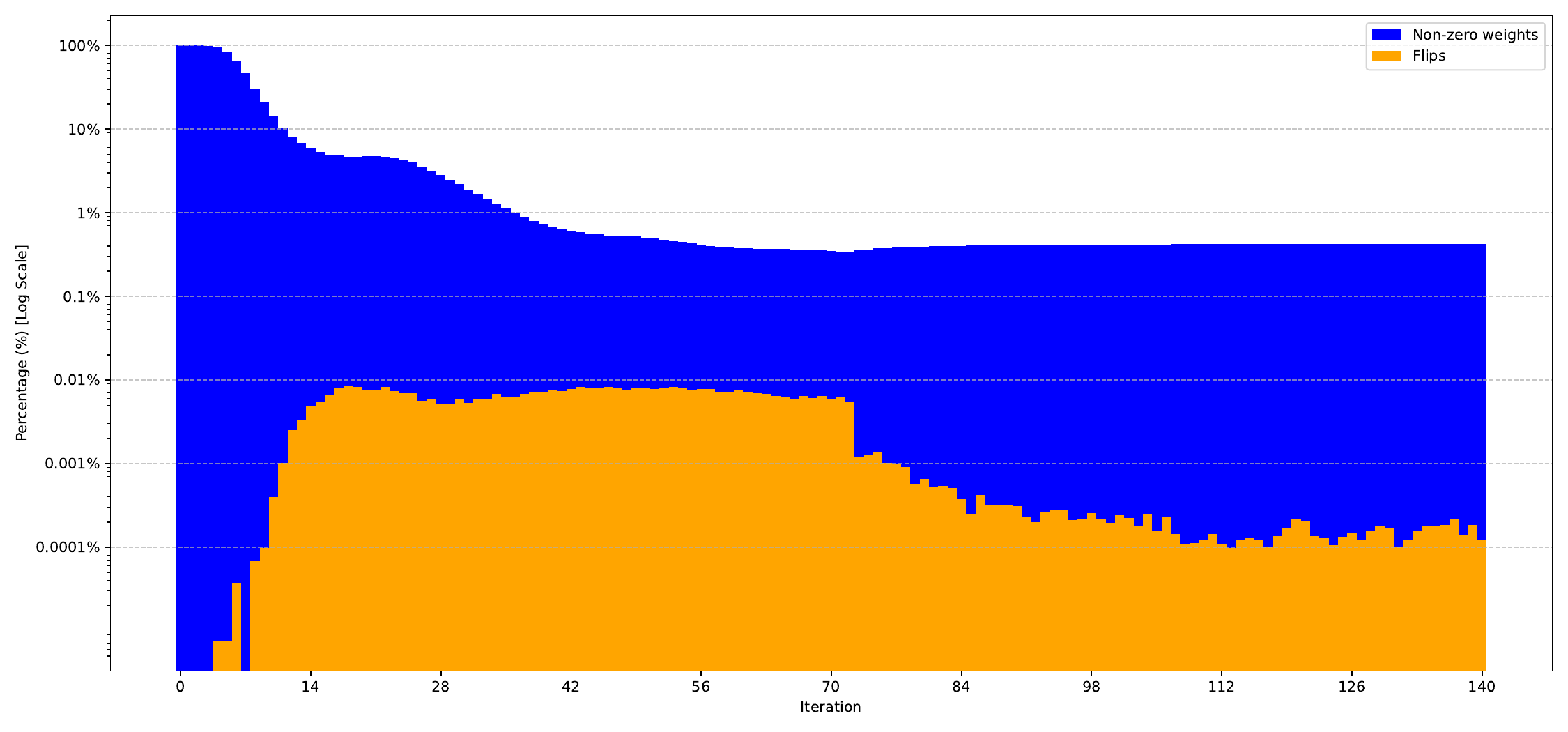}
    \caption{Frequency of Flips: The blue histogram represents the percentage of remaining parameters on a logarithmic scale, while the orange histogram illustrates the ratio of parameter flips per iteration relative to the total number of network parameters, also on a logarithmic scale.}
    \label{fig:flip_histogram}
\end{figure}

Implicit regrowth is the main source of noise in our network, leading to diverse topologies during training. In Figure~\ref{fig:flip_histogram}, we identify patterns in flip frequency, such as the lower number of flips at the start of training, which happens due to the warmup period when $t$ values drift towards the pruning threshold. 

As training progresses and the number of parameters declines, the number of flips remains relatively steady. When the scheduler is turned off and regrowth begins, a sharp decrease in the total number of flips takes place as the network stabilizes. This pattern is visible between iterations 70 and 130, alongside a gradual increase in the number of parameters.

\begin{figure}[!htbp]
  \centering
  \begin{subfigure}[b]{0.48\textwidth}
    \centering
    \includegraphics[width=\linewidth]{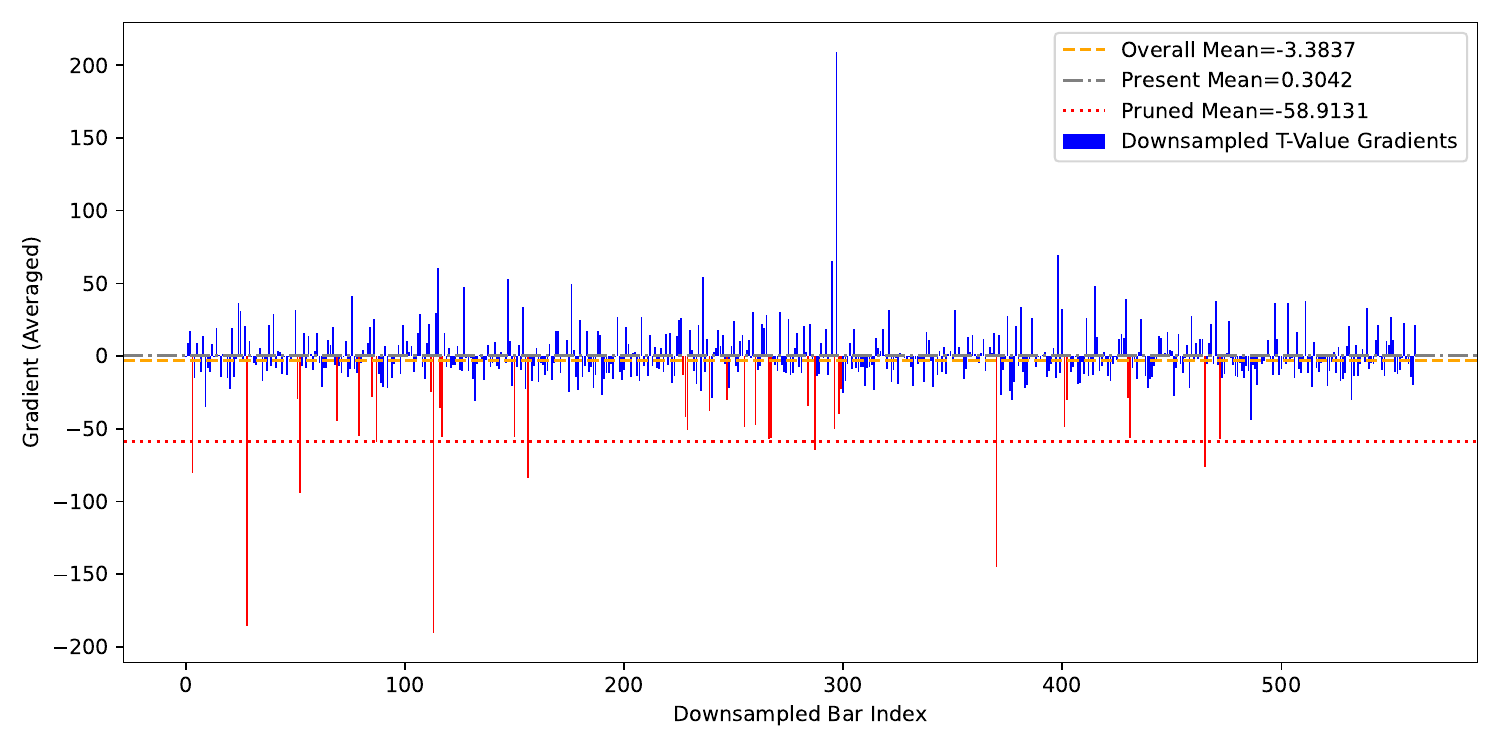}
    \caption{Weight $\omega_{14}$ gradients.}
    \label{fig:sub1_3}
  \end{subfigure}
  \hfill
  \begin{subfigure}[b]{0.48\textwidth}
    \centering
    \includegraphics[width=\linewidth]{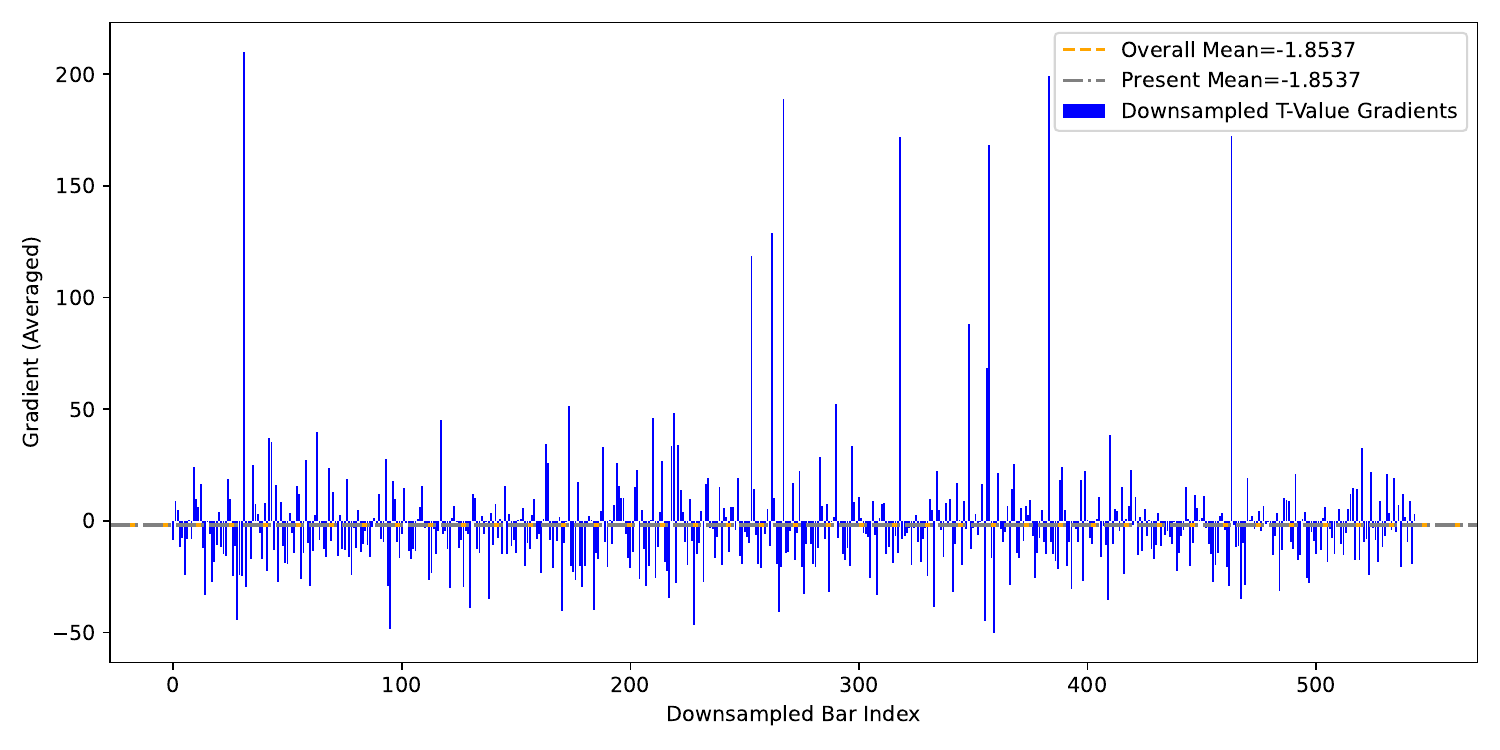}
    \caption{Weight $\omega_{13}$ gradients.}
    \label{fig:sub2_3}
  \end{subfigure}

  \vspace{1em}

  \begin{subfigure}[b]{0.48\textwidth}
    \centering
    \includegraphics[width=\linewidth]{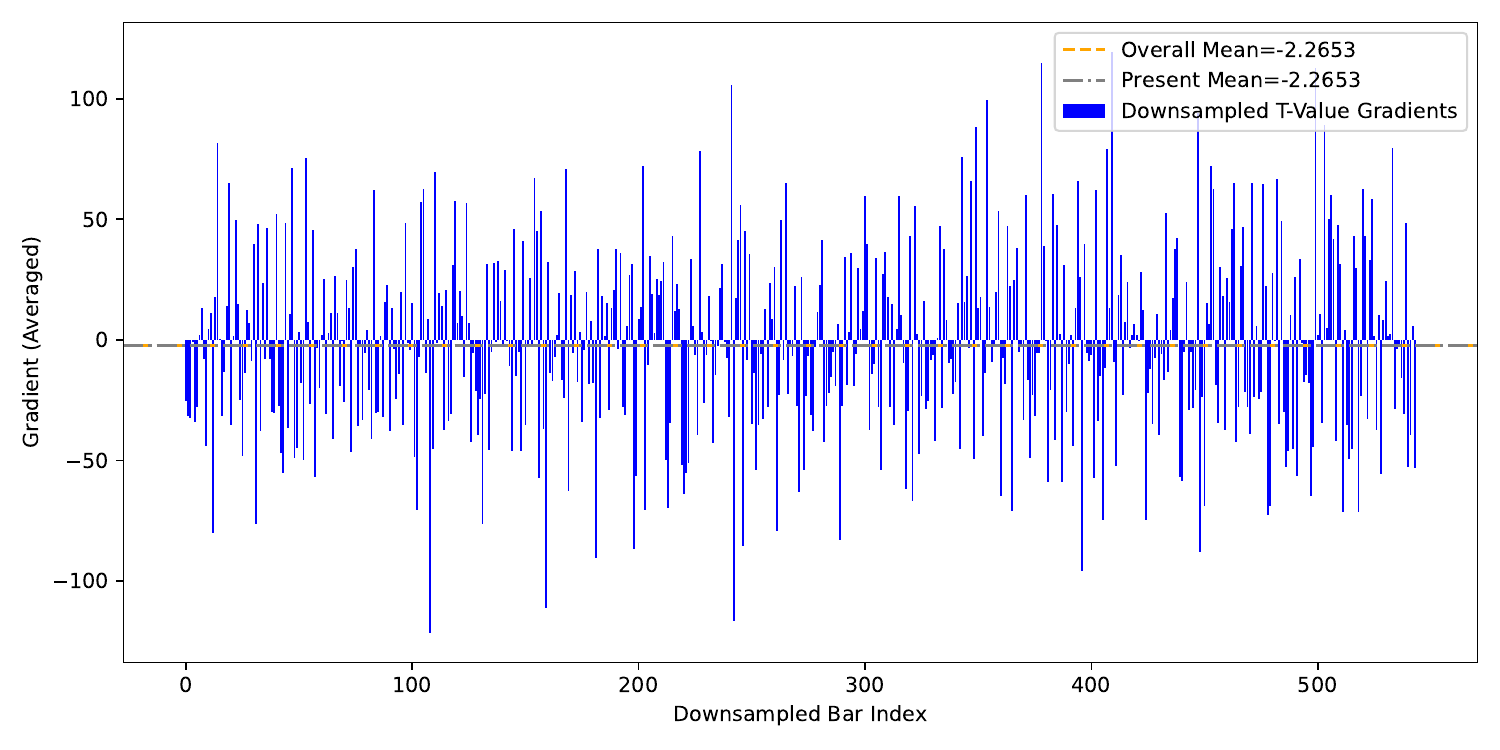}
    \caption{Weight $\omega_{11}$ gradients.}
    \label{fig:sub3_3}
  \end{subfigure}
  \hfill
  \begin{subfigure}[b]{0.48\textwidth}
    \centering
    \includegraphics[width=\linewidth]{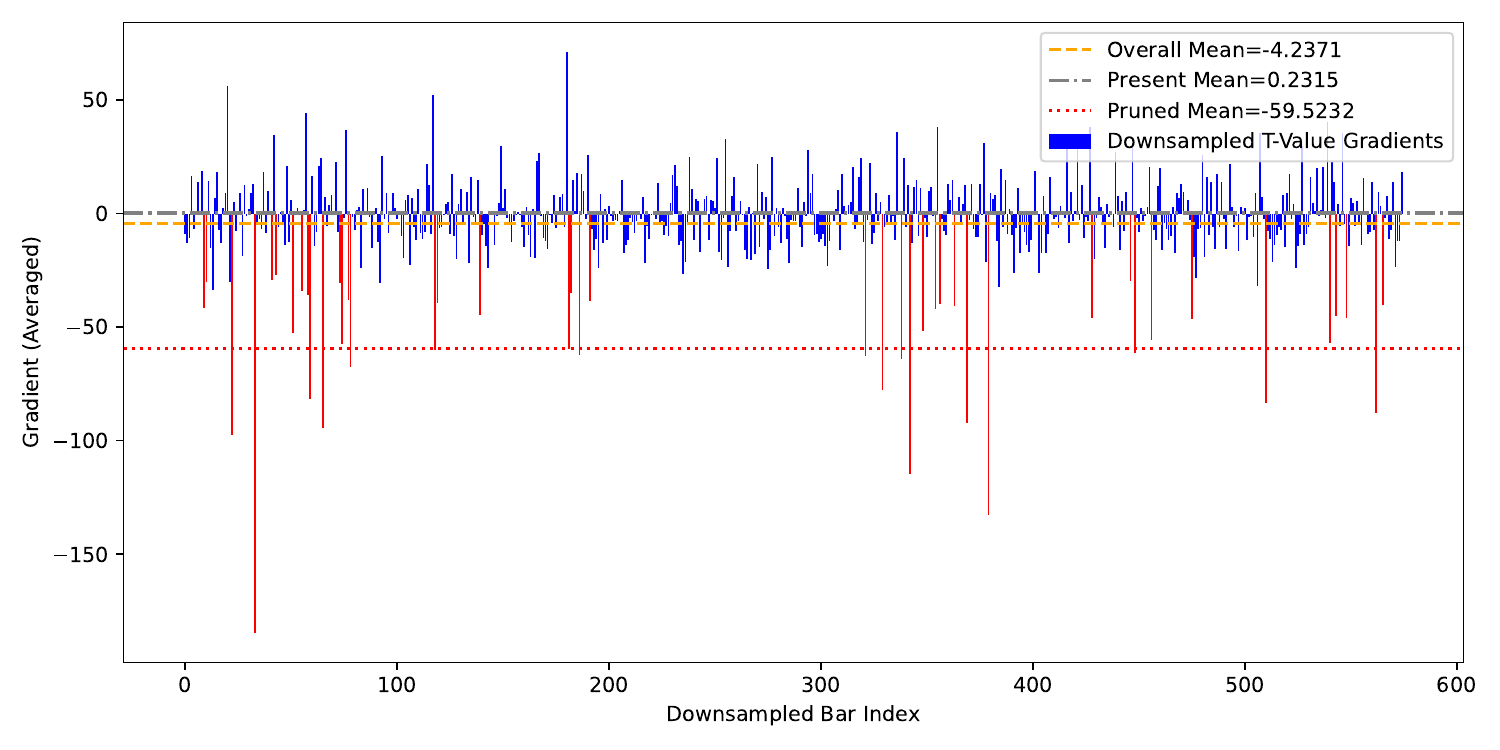}
    \caption{Weight $\omega_{12}$ gradients.}
    \label{fig:sub4_3}
  \end{subfigure}

  \caption{Gradient values over time for four remaining weights in the pruned network. Blue bars show gradients when $t_i>0$, red when $t_i\le0$. Notice the high‐magnitude red gradients ($\mathcal{G}_i^-(\omega, \mathcal{T})$) versus the typically smaller positive gradients ($\mathcal{G}_i^+(\omega, \mathcal{T})$).}
  \label{fig:ti_grad_behavior}
\end{figure}

In Figure~\ref{fig:ti_grad_behavior} we can observe the behavior of \(t\) values gradients. We analyze the gradients while the weight is pruned (pruned gradients) and present (present gradients). Recall that negative values of the gradients translate into positive updates for \(t\) values and vice-versa. (e.g., if the overall present mean is -2, then the overall change in t was 2, meaning the magnitude tends to increase). 

Two specific types of weights emerge, which confirm our findings in Section \ref{subsubsec:aggregated_flux}. In the first type (figures \ref{fig:sub1_3} and \ref{fig:sub4_3}) the present gradient has a small positive mean (on average $t_i$ decreases, \(\omega_i\) drifts towards 0, $\widehat{\mathcal{G}}_i^+ < 0$), which leads to pruning. However, these specific weights were important to the network, which creates a large flux to regrow them ($\widehat{\mathcal{G}}_i^- >> 0$), with the process repeating several times. In the second type, the overall gradient is negative (meaning $t_i$ values increase, so $\omega_i$ drifts away from 0) enough to overcome the pressure, and the weights never get pruned, which confirms one of the described weight types in Section \ref{subsubsec:aggregated_flux}. Figure \ref{fig:ti_grad_behavior} shows that, despite the interconnected dynamics of a neural network, pruning behavior in Hyperflux can still be well understood at weight level. 

\section{Complexity Analysis}
\label{appendixsec:Complexity_analysis}

We analyze the computational complexity of Hyperflux by comparing its training and inference FLOPs to a standard dense baseline. Following standard practice \citep{peste2021acdc, evci2020rigging}, we approximate complexity by counting matrix multiplications in the forward and backward passes.

\subsection{Training Complexity}

Let $f_d$ and $f_s$ denote the FLOPs required for a forward pass through a dense and sparse layer, respectively. In a standard dense network, the backward pass requires $2f_d$: one matrix multiplication to compute gradients with respect to weights ($\nabla_W L$) and one to compute gradients with respect to activations ($\nabla_a L$). Including the forward pass, the total cost is $3f_d$.

For Hyperflux, the training cost per iteration is derived as follows:
\begin{itemize}
    \item \textbf{Forward Pass:} Computed using only active weights, costing $f_s$.
    \item \textbf{Backward Pass (Weights/$t$-values):} To update the $t$-values (which govern the existence of all possible connections), we must compute gradients for the full weight matrix. This requires a dense matrix multiplication, costing $f_d$.
    \item \textbf{Backward Pass (Activations):} To propagate gradients to the preceding layer, we only require the transpose of the current sparse weight matrix. This requires a sparse matrix multiplication, costing $f_s$.
\end{itemize}

The total training cost for Hyperflux is thus:
\begin{equation}
    \text{FLOPs}_{\text{train}} = f_s + (f_d + f_s) = 2f_s + f_d
\end{equation}

The relative training cost compared to a dense baseline ($3f_d$) is:
\begin{equation}
    \text{Cost}_{\text{relative}} = \frac{2f_s + f_d}{3f_d} = \frac{2}{3}\frac{f_s}{f_d} + \frac{1}{3}
\end{equation}

\subsection{Dynamic FLOP Estimation}

As Hyperflux is a dynamic pruning method, the sparsity level $f_s$ evolves. The average relative cost over $N$ training epochs is calculated as:
\begin{equation}
    \text{Cost}_{\text{avg}} = \frac{1}{N} \sum_{i=1}^{N} \left( \frac{2}{3}\frac{f^i_s}{f_d} + \frac{1}{3} \right)
\end{equation}
where $f^i_s$ represents the sparse FLOPs at epoch $i$. Since $f_d$ is constant, this simplifies to:
\begin{equation}
    \text{Cost}_{\text{avg}} = \frac{2}{3} \frac{\sum_{i=1}^{N} f^i_s}{N \cdot f_d} + \frac{1}{3}
\end{equation}

\subsection{Inference Complexity}

During inference, the $t$-values are stationary and only the sparse weights are utilized. The complexity is identical to a standard sparse forward pass:
\begin{equation}
    \text{FLOPs}_{\text{inference}} = f_s
\end{equation}

\section{Schedulers implementation}
\label{appendixsec:schedulers-implementation}

In this section we provide two concrete implementations for the scheduler pressure policy \(\Pi\) used in our experiments. Let \(d : \mathbb{N}_0 \to (0, 1]\) be a non-increasing function mapping epoch \(e\) to a network density shrinking factor, with \(d(0) = 1\). Next, let $f : \mathbb{N}_0 \to (0, 100]$ be the density curve, defined as
$$f(e) = 100 \cdot \prod_{i=0}^{e} d(i)$$
representing the target network density (\%) at epoch $e$, with $f(0) = 100$. For example, if \(d(1) = 0.9\) and \(d(2) = 0.8\), the target density at epoch 2 will be \(f(2) = 100 \cdot 0.9 \cdot 0.8 = 72\).

\subsection{Pressure scheduler with trajectory control \(\Pi_1\) }

We denote the first policy \(\Pi_1\). Its purpose is to track a user-defined density curve at each epoch by increasing the pressure when density is above the curve (too many parameters) and decreasing pressure when density is below the curve. Concretely, the decisions at each step are taken as in Algorithm \ref{algoritmappendix:pressure-scheduler-trajectory}. This scheduler is able to follow a specific density trajectory, but its convergence density can have a relative deviation of up to 20\% from the target density. 

\begin{algorithm}[htbp]
\caption{Pressure policy \(\Pi_1\) for trajectory control}
\label{algoritmappendix:pressure-scheduler-trajectory}
\begin{algorithmic}[1]
\State \textbf{Initialization:} 
  Density function \(f\).
\State \textbf{Policy parameters:}
  Pruning epochs \(E_p\),
  Final density \(\mathcal{D}\),
  Current density \(d_e\),
  Current epoch \(e\),
  \If{\(d_e > f(e)\)}         
    \State Return true
  \Else     
    \State Return false
  \EndIf
\end{algorithmic}
\end{algorithm}

\subsection{Pressure scheduler with upper boundary}

We denote the second policy $\Pi_2$. Its purpose is to reach the final density within a relative deviation of under $5\%$ from $\mathcal{D}$, trading exact control over the sparsity curve for convergence precision. We introduce an upper boundary $ub$ to squeeze the network density between $\mathcal{D}$ and $ub$ itself. The upper boundary is defined like the density curve but recalculated at each epoch, which keeps it constant when the network density is above its target and tightens it when the density is below, thereby squeezing the network density toward $\mathcal{D}$. The algorithm for $\Pi_2$ is presented in Algorithm \ref{algoritmappendix:pressure-scheduler-upper-boundary}. Since \textit{ub} is recalculated at each epoch, accessing indexes other than 1 is meaningless. We nevertheless compute the full curve to ensure the network trajectory is steered toward $\mathcal{D}$.

\begin{algorithm}[htbp]
\caption{Pressure policy \(\Pi_2\) for upper boundary}
\label{algoritmappendix:pressure-scheduler-upper-boundary}
\begin{algorithmic}[1]
\State \textbf{Initialization:} 
  Upper boundary function \(ub\).
\State \textbf{Policy parameters:}
  Pruning epochs \(E_p\),
  Final density \(\mathcal{D}\),
  Current density \(d_e\),
  Current epoch \(e\),
\State \textbf{Internals:} 
  Density history \(dh\).
  
  \State dh.append(\(d_e\)).
  \State Recalculate \(ub\) such that \(ub(E_p-e) = \mathcal{D}\) (will reach \(\mathcal{D}\) in the remaining epochs).
  \State prev\_decrease \( \gets \) \(\frac{\text{dh}(e)}{\text{dh}(e-1)}\)
  
  \If{prev\_decrease < \( ub(1)\)}         
    \State Return true
  \Else     
    \State Return false
  \EndIf
\end{algorithmic}
\end{algorithm}

\section{Training setup and reproducibility}
\label{appendixsec:training_setup}

\begin{table}[h]
\centering
\caption{Hyperparameter configurations for the pruning and stabilization stages across
CIFAR-10, CIFAR-100, and ImageNet-1K with ResNet-50, VGG-19, and Vision Transformer
architectures.}
\label{tab:all_exp}
\small
\begin{tabular}{c | cc | cc | ccc}
\toprule
\textbf{Dataset}
  & \multicolumn{2}{c|}{\textbf{CIFAR-10}}
  & \multicolumn{2}{c|}{\textbf{CIFAR-100}}
  & \multicolumn{3}{c}{\textbf{ImageNet-1K}} \\
\midrule
\textbf{Network}
  & \textbf{ResNet-50} & \textbf{VGG-19}
  & \textbf{ResNet-50} & \textbf{VGG-19}
  & \textbf{ResNet-50} & \textbf{ViT-S/16} & \textbf{ViT-T/16} \\
\textbf{Acc (\%)}
  & 94.72 $\pm$ 0.05 & 93.85 $\pm$ 0.06
  & 78.32 $\pm$ 0.08 & 73.44 $\pm$ 0.09
  & 77.01
  & 72.20 & 78.90 \\
\midrule
\textbf{Batch size}  & 128 & 128 & 128 & 128 & 1024 & 1024 & 1024 \\
\midrule
Total epochs         & 160 & 160 & 160 & 160 & 100  & 150  & 150  \\
\(E_p / E_s\)        & 100/60 & 100/60 & 100/60 & 100/60 & 80/20 & 100/50 & 100/50 \\
\midrule
\(\mathcal{O}_t\)    & ADAM & ADAM & ADAM & ADAM & ADAM  & ADAM  & ADAM  \\
\(\mathcal{O}_\omega\) & SGD & SGD & SGD & SGD & SGD  & AdamW & AdamW \\
\(\mathcal{S}_\omega\) & Cosine & Cosine & Cosine & Cosine & Cosine & Cosine & Cosine \\
\(\lambda_\text{wd}\) & $5{\times}10^{-4}$ & $5{\times}10^{-4}$ & $5{\times}10^{-4}$ & $5{\times}10^{-4}$ & $10^{-4}$ & $0.05$ & $0.05$ \\
\midrule
\multicolumn{8}{c}{\textit{Pruning}} \\
\midrule
\(\eta_\omega^s\)    & 0.1   & 0.1   & 0.1   & 0.1   & 0.01  & $10^{-4}$ & $10^{-4}$ \\
\(\eta_\omega^e\)    & 0.003 & 0.003 & 0.003 & 0.003 & 0.003 & $10^{-5}$ & $10^{-5}$ \\
\(\eta_t\)           & 0.001 & 0.001 & 0.001 & 0.001 & 0.001 & 0.001 & 0.001 \\
\midrule
\multicolumn{8}{c}{\textit{Stabilization}} \\
\midrule
\(\eta_\omega^i\)    & 0.001  & 0.001  & 0.001  & 0.001  & 0.001  & $5{\times}10^{-4}$ & $5{\times}10^{-4}$ \\
\(\eta_\omega^f\)    & 0.0001 & 0.0001 & 0.0001 & 0.0001 & 0.0001 & $10^{-5}$ & $10^{-5}$ \\
\(\eta_t\)           & 0.001 & 0.001 & 0.001 & 0.005 & 0.003 & 0.003 & 0.003 \\
\(\mathcal{S}_t\)    & LambdaLR & LambdaLR & LambdaLR & LambdaLR & LambdaLR & LambdaLR & LambdaLR \\
\(\lambda_t\)        & 0.75 & 0.75 & 0.75 & 0.75 & 0.55 & 0.55 & 0.55 \\
\bottomrule
\end{tabular}
\end{table}

As summarized in Table~\ref{tab:all_exp}, our training protocol consists of two stages over a fixed number of epochs: a pruning stage followed immediately by a stabilization (regrowth) stage. In both stages, weights \(\omega\) are optimized by SGD under a cosine annealing scheduler \(\mathcal{S}_w\), while presence parameters \(t\) are optimized with ADAM. Furthermore, the presence parameters are uniformly initialized in the range 0.2–0.5. 

During the pruning stage, the learning rate for \(\omega\) is decayed from \(\eta_\omega^s\) to \(\eta_\omega^e\), and \(t\) uses a constant \(\eta_t\). Without resetting training, we then set the pruning pressure to zero and enter the stabilization stage, where \(\eta_\omega\) is further decayed (from its new \(\eta_\omega^i\) to \(\eta_\omega^f\)) and the presence parameters are trained under a LambdaLR scheduler \(\mathcal{S}_t\) with decay parameter \(\lambda_t\). This two-stage setup, with separate optimizer and learning rate schedules for weights and presence parameters, ensures that both the sparse structure and the remaining weights are allowed to converge to their optimal configurations. Let \(\mathcal{O}_w\) be the optimizer for the weights and \(\mathcal{O}_t\) the optimizer for the presence parameters. We prune for \(E_p\) epochs and then enter a stabilization stage lasting \(E_s\) epochs, for a total of \(E_p + E_s\) epochs. All experiments use the pressure scheduler presented in Algorithm~\ref{algoritmappendix:pressure-scheduler-trajectory}. 

Across all experiments, we applied a weight decay of $10^{-4}$, while omitting any weight decay on the batch-normalization layers. Regarding augmentations, on ImageNet we adopt the same pipeline as our baselines: random resize, crop and random horizontal flip for training, and resize plus center crop for validation; on CIFAR-10/100 we apply random crop with padding, random horizontal flip for training, and no augmentations for testing.

For the schedulers parameters, we found that a value of \(0.1\) works best for the \(u\) parameter in our scheduler for vision tasks, while for \(\alpha\) we found \(1.5\).

\end{document}